\pdfoutput=1

\documentclass[11pt]{article}

\usepackage{acl}

\usepackage{times}
\usepackage{latexsym}

\usepackage[T1]{fontenc}

\usepackage[utf8]{inputenc}

\usepackage{microtype}
\usepackage{times}
\usepackage{latexsym}
\usepackage{subcaption}
\usepackage{graphics}
\usepackage{graphicx}
\usepackage{hyperref}
\usepackage{dsfont}
\usepackage{amsmath}
\usepackage{xspace}
\usepackage{verbatim}
\usepackage{xcolor}
\usepackage{soul}
\usepackage{graphicx}
\usepackage{tabularx}
\usepackage{booktabs}
\usepackage{caption}
\usepackage{geometry}
\usepackage{algorithm}
\usepackage[noend]{algpseudocode}
\usepackage{comment}
\usepackage{todonotes}
\usepackage{subfloat}
\usepackage{float}
\usepackage{tabularray}
\usepackage{inconsolata}
\usepackage{hyperref}
\usepackage{multirow}  
\usepackage{enumitem}
\newcommand{\mavenere}{{\fontfamily{lmtt}\selectfont MAVEN-ERE} }

\title{Are LLMs Good Annotators for Discourse-level Event Relation Extraction?}

\author{Kangda Wei, Aayush Gautam, Ruihong Huang\\ 
Department of Computer Science and Engineering\\
Texas A\&M University, College Station, TX\\
\normalsize{\texttt{\{kangda, aayushgautam, huangrh\}@tamu.edu}}
}

\begin{document}
\maketitle
\begin{abstract}
Large Language Models (LLMs) have demonstrated  proficiency in a wide array of natural language processing tasks. However, its effectiveness over discourse-level event relation extraction (ERE) tasks remains unexplored. In this paper, we assess the effectiveness of LLMs in addressing discourse-level ERE tasks characterized by lengthy documents and intricate relations encompassing coreference, temporal, causal, and subevent types. Evaluation is conducted using an commercial model, GPT-3.5, and an open-source model, LLaMA-2. Our study reveals a notable underperformance of LLMs compared to the baseline established through supervised learning. Although Supervised Fine-Tuning (SFT) can improve LLMs performance, it does not scale well compared to the smaller supervised baseline model. 
Our quantitative and qualitative analysis shows that LLMs have several weaknesses when applied for extracting event relations, including a tendency to fabricate event mentions, and failures to capture transitivity rules among relations, detect long distance relations, or comprehend contexts with dense event mentions. Code available at: \url{https://github.com/WeiKangda/LLM-ERE.git}.
\end{abstract}

\vspace{0.05cm}
\section{Introduction}
\begin{figure}[t]
    \centering
    \includegraphics[scale=0.42, trim={1.5cm 0.25cm 2cm 0cm}]{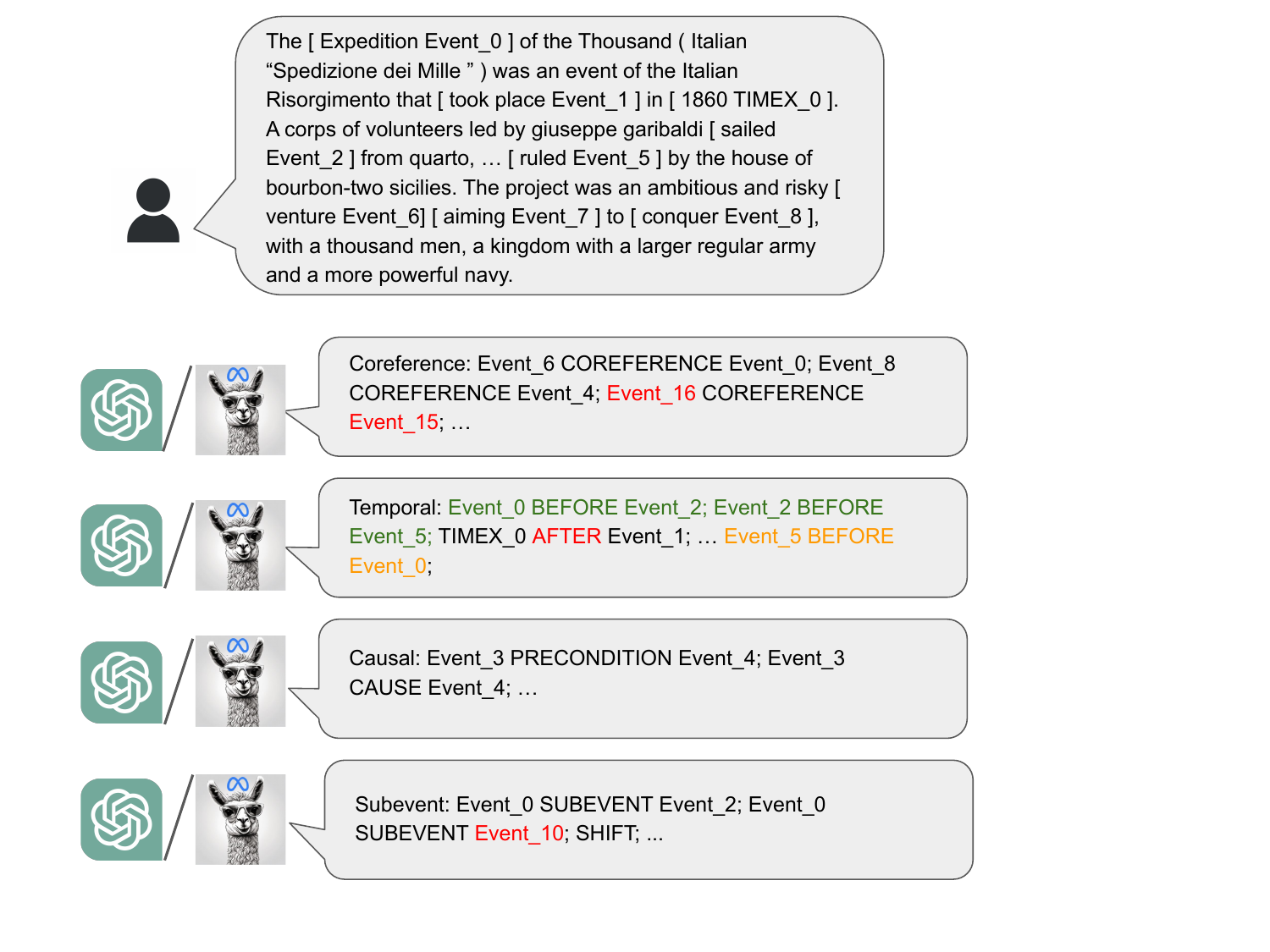}
    \vspace{-2em}
    \caption{LLMs on \mavenere with four different relation types. \mavenere is an event relation extraction dataset covering four types of relations (coreference, temporal, causal, and subevent) with long documents. Pairs in \textcolor{orange}{orange} indicate a violation of transitivity rules against pairs in \textcolor{green}{green}. Text in \textcolor{red}{red} indicates hallucinations. }
    \label{fig:intro}
\end{figure}

Event Relation Extraction (ERE) refers to the NLP tasks that  identify and classify relationships between events mentioned in a text. The commonly studied event relations include coreference, temporal, causal and subevent relations. 
ERE tasks aim to comprehend the intricate relationships between events and are beneficial for many applications, such as event prediction \cite{chaturvedi-etal-2017-story, bai-etal-2021-integrating}, question answering \cite{10.1145/3018661.3018737}, and reading comprehension \cite{berant-etal-2014-modeling}.

ERE tasks remain difficult and the empirical performance on these tasks are often rather low. 
Recently, inspired by the recent success of LLMs, \citet{yuan2023zeroshot} evaluates ChatGPT \footnote{\url{https://openai.com/blog/chatgpt}} on one ERE task, temporal relation extraction, using several benchmark datasets, including TDDiscourse \cite{naik-etal-2019-tddiscourse} that addresses temporal relation extraction at the discourse level. However, \citet{yuan2023zeroshot} has to truncate the documents and constrain the size of discourse as inputs longer than eight sentences cause ChatGPT to generate unformatted answers.
Meanwhile, \citet{gao-etal-2023-chatgpt} only evaluates ChatGPT on its causal reasoning ability. Unlike these works, we evaluate LLMs on performing multiple ERE tasks at the document level that tend to feature dense and long distance event relations, as shown in Figure~\ref{fig:intro}. Specifically, we experiment with two LLMs, the commercial model  GPT-3.5 and the open-source model LLaMA-2. 

Furthermore, humans usually benefit from comprehending the meanings of event relations through examples \cite{hu2023protoem} when conducting the ERE tasks, but the recent studies \cite{wei2023zeroshot, yuan2023zeroshot} 
often evaluate LLMs under the zero-shot setting and have a chance to limit the models in fully showcasing their true capabilities in event relation extraction. To fill the gap, we run experiments in the one-shot or few-shot settings, we also run Supervised Fine-Tuning (SFT) experiments aiming to further enhance LLMs on performing ERE tasks.

We systematically evaluate the effectiveness of LLMs on extracting four common types of event relations (coreference, temporal, causal, and subevent) using the \mavenere dataset \cite{wang-etal-2022-maven} by experimenting with multiple prompts. The design of our prompts were informed by the prior work \cite{yuan2023zeroshot, bohnet-etal-2023-coreference} to ensure the most effective prompts applied. 
Comprehensive analysis, both quantitative and qualitative, 
lead to the following findings:

\begin{itemize}[leftmargin=*] 
    \item \textbf{Performance: } Even with careful prompting, both GPT-3.5 and LLaMA-2 significantly underperform the baseline established through full supervised learning, and this is true for all the four types of relations. Although Supervised Fine-Tuning (SFT) can yield improved performance for LLaMA-2, there is still a gap between its performance and the performance of the baseline model trained with the same size of data, not to mention SFT for LLaMA-2 requires much more time and computation.
    \item \textbf{Transitivity and hallucinations: } There is a discernible tendency for LLMs to violate the rules of transitivity among event relation predictions. Furthermore, LLMs display inconsistencies in adhering to the provided prompts. 
    Both suggest a potential lack of reasoning abilities and inaccurate understanding on the assigned task.
    \item \textbf{Events Distance and Density: } LLMs encounter challenges in capturing 
    long distance event relations and inter-sentence event relations. LLMs also struggle to capture event relations in complex contexts that are dense with event mentions.
\end{itemize}

\section{Related Works}

\paragraph{Event Relation Extraction}
Event relation extraction (ERE) has been one of the fundamental challenges for natural language processing \cite{chaturvedi-etal-2017-story, rashkin-etal-2018-event2mind, ijcai2020p554}. As understanding relations between events is crucial for understanding human languages \cite{Levelt89, MillerJohnsonLaird+1976} and 
beneficial for various applications \cite{khashabi2019question, ijcai2020p554, choubey-huang-2018-improving}, many approaches have been developed for performing ERE tasks
\cite{liu-etal-2014-supervised, hashimoto-etal-2014-toward, ning-etal-2017-structured}, and many high performing approaches are based on supervised learning \cite{dligach-etal-2017-neural, aldawsari-finlayson-2019-detecting, ijcai2020p499, lu-ng-2021-constrained}.

\paragraph{LLMs for Extraction Tasks}
LLMs have been applied to several common information extraction tasks including event extraction, 
relation extraction and named entity recognition \cite{gonzálezgallardo2023yes, borji2023categorical, tang2023does, gao2023exploring, wei2023zeroshot}. However, to the best of our knowledge, LLMs have not been well explored for ERE tasks. 
Recently, \citet{yuan2023zeroshot} evaluates ChatGPT on temporal relation extraction and \citet{gao-etal-2023-chatgpt} evaluates ChatGPT on causal reasoning with the binary Event Causal Identification task, in contrast, we evaluate LLMs on extracting multiple types of fine-grained event relations. 
The dataset we use in this study, \mavenere \cite{wang-etal-2022-maven}, has dense relations at the discourse-level for four common types of event relations: coreference, temporal, causal, and subevent.

\paragraph{Prompt Engineering}
Many recent works have studied how to make LLMs perform better through applying various prompting techniques, including role-prompting \cite{Zhang_2023, vanburen2023guided}, re-sampling \cite{holtzman2020curious}, one-shot or few-shot prompting \cite{logan2021cutting, shyr2023identifying}, and question decomposition \cite{wei-etal-2023-decompositions}. Other novel and advanced techniques include Chain of Thought prompting \cite{wei2023chainofthought}, least-to-most prompting \cite{zhou2023leasttomost}, learning from multiple prompts \cite{wei-etal-2023-leveraging}, and retrieval augmentation \cite{lazaridou2022internetaugmented, jiang2023active}. 
We refer to these techniques as guidelines when designing prompts in this work.

\section{Experiment Setup}

\subsection{Data}
For this study, we use the \mavenere dataset created by \citet{wang-etal-2022-maven}, which includes annotations of four types of event relations: coreference, temporal, causal, and subevent. \mavenere consists of 4480 English Wikipedia documents, containing $103,193$ event coreference chains, $1,216,217$ temporal relations, $57,992$ causal relations, and $15,841$ subevent relations. 

\mavenere is challenging as the documents contain comprehensive relation types, event relations at the discourse level and have denser relations among events comparing to other datasets. For example, \mavenere has an average of 272 temporal relation links per document comparing to 49 temporal relation links per document for MATRES \cite{ning-etal-2018-multi}; TimeBank-Dense \cite{cassidy-etal-2014-annotation} mainly focus on sentence-level relations; and TDDiscourse \cite{naik-etal-2019-tddiscourse} only consider temporal relations. 

As we test LLMs in a prompting setting without extra fine-tuning, we only utilize ten documents from the training set to extract examples included in a prompt, 
and we use ten documents from the validation set for prompt design and 
selection. We report the performance of LLMs on the whole test set of \mavenere, which contains 857 documents with $18,908$ event coreference chains, $234,844$ temporal relations, $11,978$ causal relations, and $3,822$ subevent relations.

\subsection{Prompts}
\label{sec: prompting}

There are many possible ways to prompt LLMs for \mavenere. We design four different prompt patterns, namely Bulk Prediction, Iterative Prediction, Event Ranking referring to previous works \cite{yuan2023zeroshot, bohnet-etal-2023-coreference}, and Pairwise. In the following sections, we describe each prompt pattern. Examples of prompt patterns can be found in Appendix~\ref{sec:prompts patterns}. 

For all prompt patterns, we first label all event mentions as $[\;x_i\;$ Event$\_p]$ where $x_i$ is the triggering word in sentence $S$ and Event$\_p$ is the Event ID given a document $D = \{S_1, S_2, ..., S_n\}$. TIMEX mentions are also considered for forming temporal relations, therefore, we label TIMEX mentions as $[\;x_i\;$ TIMEX$\_q]$ and TIMEX$\_q$ is the TIMEX ID. $p$ and $q$ starts from $0$ and gets increased by 1 each time a new event mention or TIMEX mention is labeled. We define $E$ to be the set of event mentions and $T$ to be the set of TIMEX mentions. We define $Y$ to be the set of four relation types where $Y = \{$coreference, temporal, causal, subevent$\}$. $R_y$ is defined to be the set containing all the sub-relation types for $y \in Y$, where $R_{coreference} = \{$COREFERENCE$\}$, $R_{temporal} = \{$BEFORE, CONTAINS, OVERLAP, BEGINS-ON, ENDS-ON, SIMULTANEOUS$\}$, $R_{causal} = \{$CAUSE, PRECONDITION$\}$, and $R_{subevent} = \{$SUBEVENT$\}$.

\subsubsection{Bulk Prediction} 
Using the Bulk Prediction prompting, we query the LLM four times for each test document, with each query asking LLMs to list all relation pairs for each $y \in Y$. For each query, we also provide an example document followed by the gold relation pairs for the same $y$  as the query. Notice this is a 1-shot setting since we provide a whole document as an example.

\subsubsection{Iterative Prediction}
Algorithm~\ref{alg:iterative prediction} sketches the Iterative Prediction prompting method. We query LLMs by iterating through the document $D$ sentence by sentence. For each new sentence $S$, we append $S$ to all the previous sentences that are already augmented with event relations predicted by the model.
Each $S$ is queried four times for each $y \in Y$. For coreference relation, we follow the Link-Append approach proposed by \citet{bohnet-etal-2023-coreference} to augment the queried sentences. For temporal, causal, and subevent relations, we augment the sentences 
by inserting predicted relation tuples after the Event ID or TIMEX ID. A tuple is in the form $(e||t, r_y, e||t)$, where $e\in E$, $t \in T$, and $r_y \in R_y$. 

We experiment with two ways for providing demonstrations to the model: 
(1) whole doc, and (2) $n$-shot. 
\begin{algorithm}
\small
  \caption{Iterative Prediction}\label{alg:iterative prediction}
  \hspace*{\algorithmicindent} \textbf{Inputs:} Test Document $D$, Example Document $D^{'}$
  \begin{algorithmic}[1]
      \For {$y \in Y$}
        \For {$S_i \in D$}
            \If {$i == 0$}
                \State Show an example with $D^{'}$;
                \State Query LLMs to list all relation tuples occurred in $S_0$ with relation $r_y \in R_y$;
                \State Augment $S_0$ with predicted tuples;
            \Else
                \State Show an example with $D^{'}$;
                \State Append $S_i$ to augmented $S_{0:i}$
                \State Query LLMs to list all relation tuples occurred in $S_{0:i}$ with relation $r_y \in R_y$;
                \State Augment $S_i$ with predicted tuples;
            \EndIf
        \EndFor
    \EndFor
  \end{algorithmic}
\end{algorithm}

\paragraph{Whole Doc} 
For each query, we provide one full training document augmented with gold event relations as an example. The length of the example increases as we iteratively go through the document sentence by sentence. 
Eventually, the  model has access to the full example document augmented with gold relations for reference.

\paragraph{$n$-Shot} For each query, we show $n$ short documents augmented with gold relation labels as examples. Different from the Whole Doc approach described above, each document example here only consists of the first two sentences of an original training document. We retain the first two sentences of a document instead of only one sentence so that LLMs have access to event relations involving event pairs both within the same sentence and across different sentences.

Notice that the whole doc prompt utilize a whole document while the n-shot prompt only uses the first two sentences of a documents. We would like to provide whole documents for the n-shot prompt as demonstrations. However, the prompt length is constrained since there is a limitation for context length, which only allows one full document to be provided as demonstration. We designed the n-shot prompt since we would also like to explore the effects to LLMs by providing varying amounts of supervision. 

\begin{figure*}
     \centering
     \begin{subfigure}[b]{0.49\textwidth}
         \centering
         \includegraphics[width=0.85\linewidth,trim={1.5cm 0.5cm 2cm 0cm}]{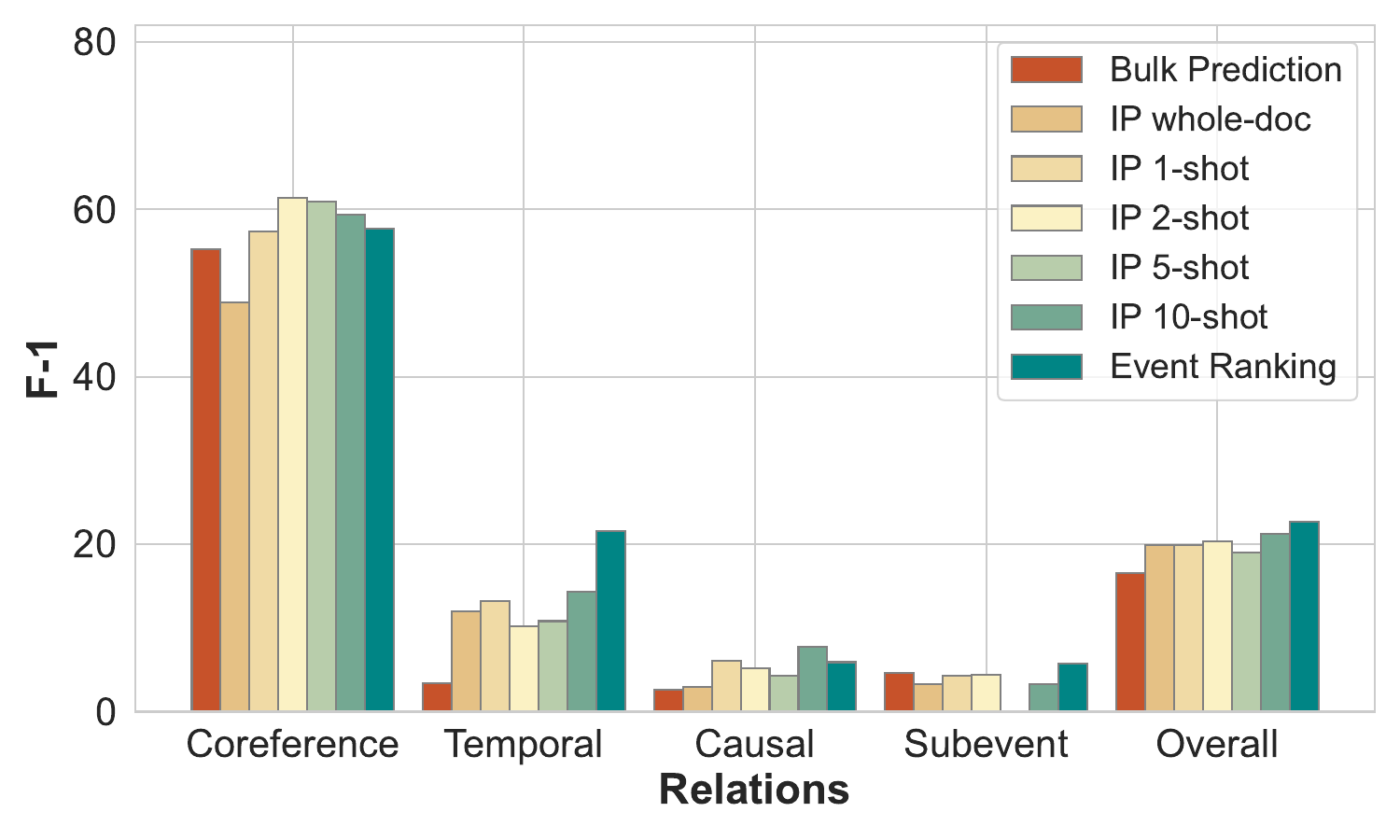}
         \caption{GPT-3.5}
         \label{fig:gpt_valid}
     \end{subfigure}
     \begin{subfigure}[b]{0.49\textwidth}
         \centering
         \includegraphics[width=0.85\linewidth,trim={1.5cm 0.5cm 2cm 0cm}]{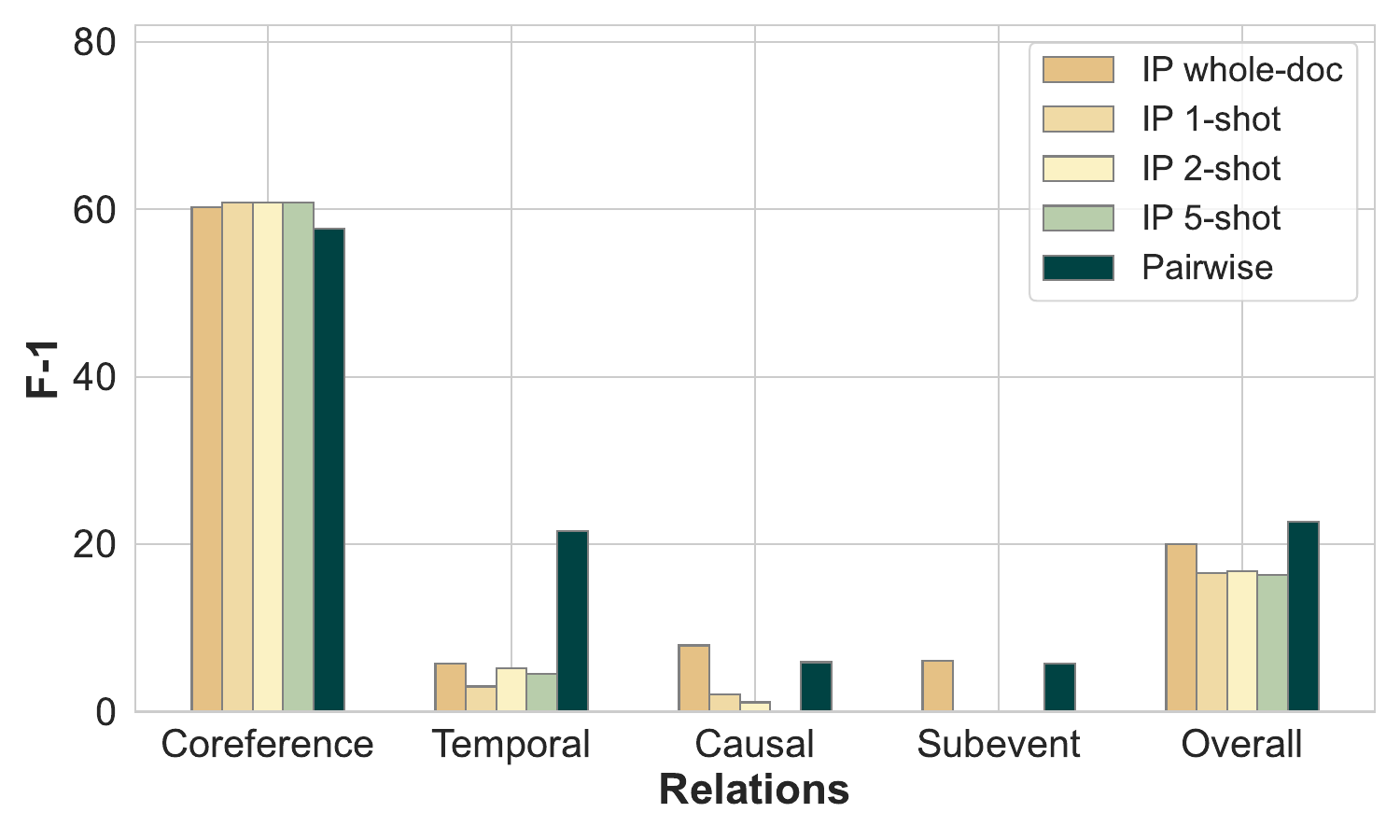}
         \caption{LLaMA-2}
         \label{fig:llama_valid}
     \end{subfigure}
        \caption{Comparisons of F-1 scores using different prompts on each type of event relations and the overall (macro average) performance, based on the results of GPT-3.5 (a) and LLaMA-2 (b) on the first 10 validation documents. The F-1 scores were calculated using the evaluation script provided by \citet{wang-etal-2022-maven}, and the detailed results of precision and recall can be found in Appendix~\ref{sec: detailed valid tables}.}
        \label{fig:main}
\end{figure*}

\subsubsection{Event Ranking} 
For the Event Ranking prompting method, we query LLMs by iterating through $e$ and $t$ for $\forall e \in E, \forall t \in T$ in test document $D$ as shown in Algorithm~\ref{alg:event ranking}. We ask LLMs to complete the query $(?, r_y, e||t)$, $\forall e \in E, \forall t \in T, \forall r_y \in R_y$, and $\forall y \in Y$. 
Note that we only need to query TIMEX mentions for temporal relation since TIMEX mentions are only relevant to temporal relations. 
\begin{algorithm}
    \small
  \caption{Event Ranking}\label{alg:event ranking}
  \hspace*{\algorithmicindent} \textbf{Inputs:} Test Document $D$, Example Document $D^{'}$
  \begin{algorithmic}[1]
  \For {$y \in Y$}
    \If {$y\;==$ temporal}
        \For {$e \in E$}
            \For {$r \in R_{y}$}
            \State Show an example with $D'$;
                \State Query LLMs with$(?, r, e)$ for $D$;
            \EndFor   
        \EndFor
        \For {$t \in T$}
            \For {$r \in R_{y}$} 
                \State Show an example with $D{'}$;
                \State Query LLMs with$(?, r, t)$ for $D$;
            \EndFor   
        \EndFor
    \Else
        \For {$e \in E$}
            \For {$r \in R_{coreference}$}
                \State Show an example with $D{'}$;
                \State Query LLMs with$(?, r, e)$ for $D$;
            \EndFor   
        \EndFor
    \EndIf
  \EndFor
\end{algorithmic}
\end{algorithm}
We also provide one example for each query. The example contains an example document, a query $(?, r_y, e^{'}||t^{'})$, where $e^{'}||t^{'}$ is an Event mention or a TIMEX mention from the example document, and the gold relation tuples as the answer. Notice the query for the test document, $(?, r_y, e||t)$, and the query in the example, $(?, r_y, e^{'}||t^{'})$, have the same $r_y$. This is also a 1-shot setting since we provide a whole document as an example.

\subsubsection{Pairwise}
\label{sec:pairwise}
For the pairwise prompting method, we query LLMs with all the event mentions and TIMEX mentions pairs. We ask LLMs to complete the query $(e||t, R_y=?, e||t)$, $\forall e \in E, \forall t \in T$, and $\forall y \in Y$. Note that we only need to query TIMEX mentions for temporal relation since TIMEX mentions are only relevant to temporal relations. If there is no relation between two events then NONE should be predicted. We also provide one example for each sub-relation $r_y$ of all four relation types. Note that this prompt pattern is in purely natural language format. However, since the number of event pairs equals the number of times we query LLMs, which grows quickly as the number of events increases, this approach is not feasible to use for GPT-3.5 and GPT-4 if we take into account of financial and time costs. Therefore, we only test this prompt with LLaMA-2.

\subsection{Model}
For this study, we use both open-source LLMs and closed-sourced LLMs for evaluation. For open-source models, we use \textit{LLaMA-2 7B} model, specifically \textit{Llama-2-7b-chat-hf}, accessed through Huggingface\footnote{\url{https://huggingface.co/meta-llama/Llama-2-7b-chat-hf}}. For closed-source models, we consider 
the \textit{gpt-3.5-turbo-16k} model \footnote{More detailed information can be found in Appendix~\ref{sec:model versions}} from OpenAI API \footnote{\url{https://platform.openai.com/docs/models}} as ChatGPT has been the most successful commercial LLMs so far
. 
We test \textit{Llama-2-7b-chat-hf} and \textit{gpt-3.5-turbo-16k} on both validation and test sets using various different prompts.

To get an idea of how the newest GPT model performs, we also tested \textit{gpt-4-1106-preview} model on a subset of the validation set instead of the whole test set becuase API calls to GPT-4 models are 30 times more expensive than GPT-3.5 models. The model performance was reported in Appendix~\ref{sec: detailed valid tables}.

\subsection{Prompt Decision}

\label{sec:prompt decision}
Before experimenting on the whole test set containing 857 documents, we test the four prompt patterns on the first 10 documents of the \mavenere validation set as running through the entire test set is both financially and time wise expensive. Table~\ref{tab:cost_table} shows our estimates of time and money needed to run each model through the whole test set using each of the latter three relatively costly prompts. 

Figure~\ref{fig:main} shows the results of GPT-3.5 and LLaMA-2 using different prompts on the first 10 validation documents\footnote{The results of GPT-4 on the first 10 validation documents are included in Appendix~\ref{sec: detailed valid tables}.}.  
\begin{table}[t]
\small
\begin{center}
\scalebox{0.9}{\begin{tabular}{c c c c}
    \toprule
     Model & Prompt & hours & USD(\$)\\
     \midrule
     \multirow{3}{*}{GPT-3.5} & Iterative Prediction & $48$ & $300$ \\
     & Event Ranking & $600$ & $3,500$ \\
     & Pairwise & $3,600$ & $30,000$ \\
     \midrule
     \multirow{3}{*}{LLaMA-2} & Iterative Prediction & $36$ & $-$ \\
     & Event Ranking & $480$ & $-$ \\
     & Pairwise & $3,000$ & $-$ \\
     \bottomrule
\end{tabular}}
\end{center}
\vspace{-0.1in}
\caption{Estimated costs.}
\vspace{-0.in}
\label{tab:cost_table}
\end{table}
The left sub-figure shows the results of GPT-3.5. 
We do not consider the Pairwise prompt for GPT-3.5 due to the extremely high cost on money. Among the remaining three prompts, Event Ranking achieves the best overall performance and Iterative Prediction 
performs relatively close. If we look into their performance over the relation types, Iterative Prediction wins on coference and causal relations while Event Ranking wins on the other two types of relations. Considering that Event Ranking is over ten times more expensive than Iterative Prediction (table \ref{tab:cost_table}), we choose Iterative Prediction over Event Ranking for running our full experiments.  

The right sub-figure of Figure~\ref{fig:main} shows the results of LLaMA-2. LLaMA-2 failed to generate consistent outputs when using Bulk Prediction and Event Ranking, thus we do not include those numbers. Between the Pairwise prompt and Iterative Prediction\footnote{IP 10-shot was not performed for LLaMA-2 due to context length constraint.}, the Pairwise prompt yields a slightly better overall performance. However, across the four relation types, the Iterative Prediction prompt, IP whole-doc, wins on three relation types (coreference, causal and subevent relations) while the Pairwise prompt only wins on temporal relations. In addition to performance, we also consider the dramatic running time differences between the two prompts (table \ref{tab:cost_table}), and we choose Iterative Prediction over the Pairwise prompt for running full experiments using LLaMA-2. 

To summarize, we choose Iterative Prediction with its variations as the final prompts for both models when running the full experiments. 

\subsection{Supervised Baseline}
We consider the baseline model proposed by \citet{wang-etal-2022-maven} as our baseline model. The baseline model utilizes \textit{roberta-base} model from Huggingface\footnote{\url{https://huggingface.co/roberta-base}} as the underlying language model and trains separate classification heads for each relation type $y \in Y$. The baseline model is trained end-to-end and performs pair-wise classification.
We train the baseline model for each relation type separately for fair comparison against GPT-3.5, and LLaMA-2. We strictly follow the training and evaluating processes according to \citet{wang-etal-2022-maven}. 

\section{Results}
\label{sec:results}
\paragraph{LLMs Performance}
\begin{table*}[t!]
\small
\footnotesize
\begin{center}
\scalebox{0.80}{\begin{tabular}{ lrrr|rrr|rrr|rrr} 
 \toprule
 Iterative& \multicolumn{3}{c|}{\textbf{MUC} \cite{vilain-etal-1995-model}} & \multicolumn{3}{c|}{\textbf{B$^{3}$} \cite{bagga-baldwin-1998-entity-based}} & \multicolumn{3}{c|}{\textbf{CEAF$_{e}$} \cite{luo-2005-coreference}} & \multicolumn{3}{c}{\textbf{BLANC} \cite{RECASENS_HOVY_2011}} \\
 Prediction & Precision & Recall & F-1 & Precision & Recall & F-1 & Precision & Recall & F-1 & Precision & Recall & F-1\\
 \midrule
 GPT-3.5 & & & & & & & & & & & &\\
 whole doc & $\mathbf{21.6}$ & $\mathbf{25.7}$ & $\mathbf{23.2}$ & $91.7$ & $\mathbf{93.2}$ & $92.5$ & $91.6$ & $89.9$ & $90.1$ & $\mathbf{57.8}$ & $\mathbf{56.3}$ & $\mathbf{56.9}$\\ 
 1-shot & $15.3$ & $17.0$ & $16.1$ & $92.0$ & $92.8$ & $92.4$ & $91.0$ & $90.1$ & $90.6$ & $54.3$ & $53.8$ & $54.0$\\ 
 2-shot & $17.9$ & $18.9$ & $18.4$ & $92.6$ & $92.9$ & $92.7$ & $91.2$ & $90.7$ & $91.0$ & $54.9$ & $54.3$ & $54.5$\\ 
 5-shot & $17.7$ & $15.2$ & $16.4$ & $\mathbf{93.9}$ & $92.7$ & $\mathbf{93.3}$ & $\mathbf{91.9}$ & $\mathbf{92.3}$ & $\mathbf{91.7}$ & $55.4$ & $53.3$ & $54.0$\\ 
 10-shot & $11.5$ & $12.0$ & $11.8$ & $92.3$ & $92.5$ & $92.4$ & $90.5$ & $90.2$ & $90.4$ & $53.2$ & $52.2$ & $52.6$\\
 \midrule
 LLaMA-2 & & & & & & & & & & & &\\
 whole doc & $\mathbf{10.6}$ & $\mathbf{6.9}$ & $\mathbf{8.4}$ & $95.1$ & $\mathbf{92.2}$ & $93.7$ & $\mathbf{90.6}$ & $93.4$ & $92.0$ & $\mathbf{53.0}$ & $\mathbf{51.1}$ & $\mathbf{51.5}$\\ 
 1-shot & $0$ & $0$ & $0$ & $\mathbf{100.0}$ & $92.0$ & $\mathbf{95.8}$ & $90.5$ & $\mathbf{98.4}$ & $\mathbf{94.3}$ & $49.3$ & $50.0$ & $49.7$\\ 
 2-shot & $0$ & $0$ & $0$ & $100.0$ & $92.0$ & $95.8$ & $90.5$ & $98.4$ & $94.3$ & $49.3$ & $50.0$ & $49.7$\\
 5-shot & $0$ & $0$ & $0$ & $100.0$ & $92.0$ & $95.8$ & $90.5$ & $98.4$ & $94.3$ & $49.3$ & $50.0$ & $49.7$\\
 \midrule
 \textbf{Baseline} & $79.8_{1.6}$ & $83.6_{0.5}$ & $81.7_{0.7}$ & $97.8_{0.2}$ & $98.4_{0.0}$ & $98.1_{0.1}$ & $98.0_{0.1}$ & $97.6_{0.2}$ & $97.8_{0.1}$ & $87.6_{1.1}$ & $92.1_{0.1}$ & $89.7_{0.6}$\\
\bottomrule
\end{tabular}}
\end{center}
\begin{center}
\scalebox{0.85}{\begin{tabular}{ lrrr|rrr|rrr|r} 
 \toprule
 Iterative& \multicolumn{3}{c|}{\textbf{Temporal}} & \multicolumn{3}{c|}{\textbf{Causal}} & \multicolumn{3}{c|}{\textbf{Subevent}} & \multicolumn{1}{c}{\textbf{Overall}} \\
 Prediction & Precision & Recall & F-1 & Precision & Recall & F-1 & Precision & Recall & F-1 & F-1\\
 \midrule
 GPT-3.5 & & & & & & & & & &\\
 whole doc & $19.8$ & $4.4$ & $7.2$ & $2.9$ & $2.9$ & $2.8$ & $\mathbf{1.9}$ & $1.3$ & $1.6$ & $19.3$\\ 
 1-shot & $17.1$ & $4.5$ & $7.1$ & $4.1$ & $2.7$ & $3.3$ & $1.9$ & $1.2$ & $1.5$ & $18.8$\\ 
 2-shot & $19.5$ & $4.3$ & $7.1$ & $4.1$ & $2.6$ & $3.2$ & $1.5$ & $0.9$ & $1.2$ & $18.9$\\ 
 5-shot & $21.3$ & $5.8$ & $9.1$ & $4.5$ & $3.0$ & $3.6$ & $1.7$ & $1.4$ & $1.6$ & $19.5$\\ 
 10-shot & $\mathbf{26.8}$ & $\mathbf{8.0}$ & $\mathbf{12.3}$ & $\mathbf{5.3}$ & $\mathbf{5.3}$ & $\mathbf{5.3}$ & $1.7$ & $\mathbf{2.8}$ & $\mathbf{2.1}$ & $\mathbf{20.4}$\\
 \midrule
 LLaMA-2 & & & & & & & & & &\\
 whole doc & $17.2$ & $\mathbf{3.1}$ & $\mathbf{5.2}$ & $4.1$ & $\mathbf{5.0}$ & $\mathbf{4.5}$ & $3.4$ & $\mathbf{6.3}$ & $\mathbf{4.4}$ & $\mathbf{18.9}$\\ 
 1-shot & $15.4$ & $1.2$ & $2.2$  & $4.6$ & $0.2$ & $0.3$ & $3.3$ & $0.1$ & $0.2$& $15.7$\\ 
 2-shot & $\mathbf{26.3}$ & $2.2$ & $4.1$ & $4.6$ & $0.1$ & $0.2$ & $0$ & $0$ & $0$ & $16.1$\\
 5-shot & $19.4$ & $1.3$ & $2.4$ & $\mathbf{8.2}$ & $0.2$ & $0.3$ & $\mathbf{4.5}$ & $0.2$ & $0.4$ & $15.8$\\
 \midrule
 \textbf{Baseline} & $57.3_{0.6}$ & $54.5_{0.1}$ & $55.8_{0.2}$ & $34.2_{0.1}$ & $29.3_{1.0}$ & $31.6_{0.6}$ & $29.5_{2.5}$ & $25.4_{2.6}$ & $27.2_{0.9}$ & $51.6$\\
\bottomrule
\end{tabular}}
\end{center}
\caption{Event coreference,temporal, causal, and subevent relations performances of GPT-3.5 and LLaMA-2 on \mavenere test set using Iterative prediction prompt patterns comparing to the baseline. The numbers in \textbf{bold} indicate the best performance across different prompt patterns for GPT-3.5 or LLaMA-2 separately. We report averages and standard deviations over 3 random seeds for the baseline method.}
\label{tab:main-table}
\end{table*}
We report the performance of GPT-3.5 and LLaMA-2 on extracting coreference, temporal, subevent, and causal relations in Table~\ref{tab:main-table}. We can see that both models severely underperform the supervised baseline model in extracting each of  the four types of relations. 

For coreference relations, we report results using four metrics following previous work \cite{wang-etal-2022-maven}.   Table~\ref{tab:main-table} shows that among the IP prompt variants, the whole doc prompt yields the best coreference resolution performance for both GPT-3.5 and LLaMA-2. For the $n$-shot prompts, the performance decreases as $n$ increases. It appears that by providing LLMs an example document with all the mentions and gold cluster included, the whole doc prompt hints LLMs to search through a whole document and better link coreferential mentions compared to the $n$-shot prompts that only provide 
excerpts from different documents.

Regarding temporal, causal, and subevent relations, GPT-3.5's performance increases as the number of example documents increases for $n$-shot prompt with the 10-shot prompt yielding the best performance. While for LLaMa-2, the whole doc prompt yields the best performance.

Regarding individual types of relations, both models have the best performance on temporal relations and the worst performance on subevent relations. We believe such a performance gap is primarily because temporal relations are much denser than subevent relations in \mavenere. 

The last column of Table~\ref{tab:main-table} shows the macro-average performance over the four types of relations. We can see that the 10-shot prompt performs slightly better than the whole doc prompt for GPT-3.5 while the whole doc prompt achieves the best performance for LLaMA-2. 

\begin{figure*}[!h]
    {\centering
    \begin{tabular}{cccc}
        \hspace{-0.3cm}\includegraphics[width=0.34\textwidth]{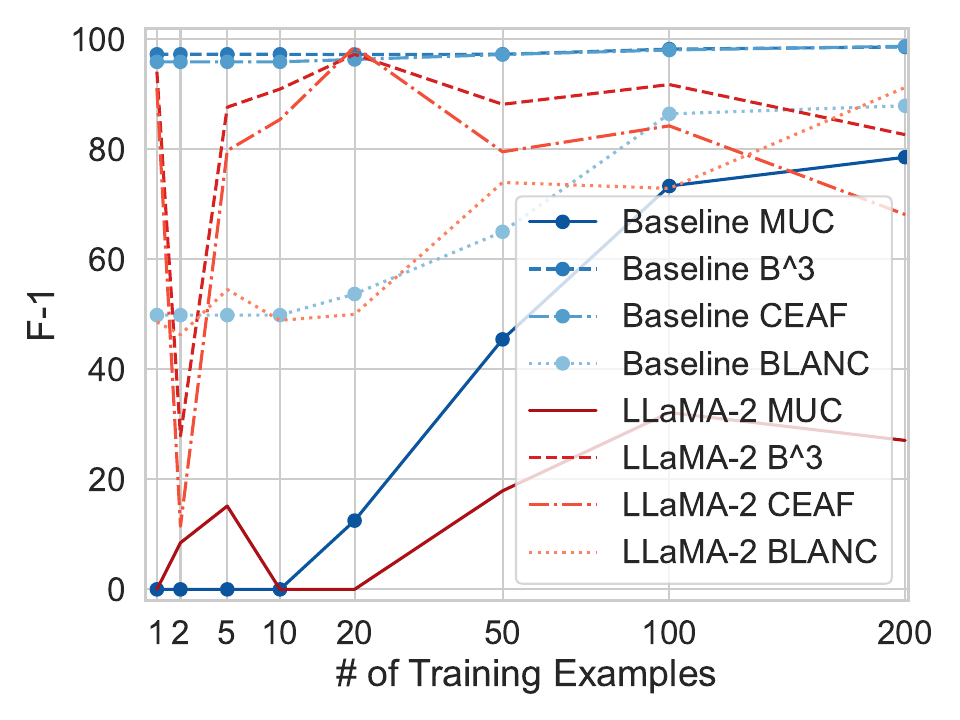}\hspace{-0.5cm} &
        \includegraphics[width=0.34\textwidth]{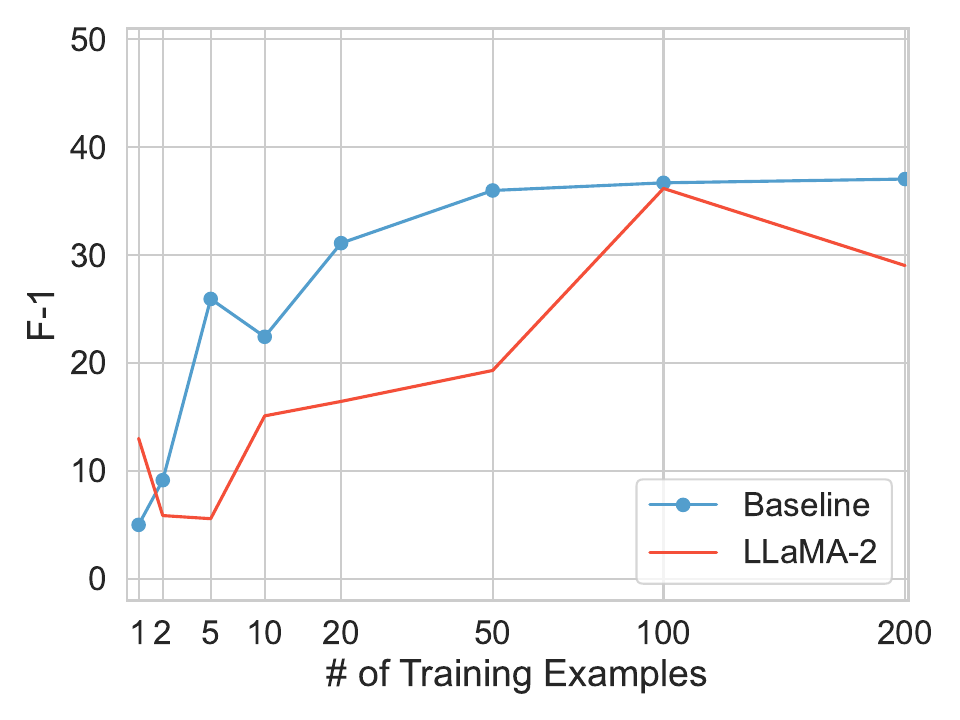}\hspace{-0.5cm} &
        \includegraphics[width=0.34\textwidth]{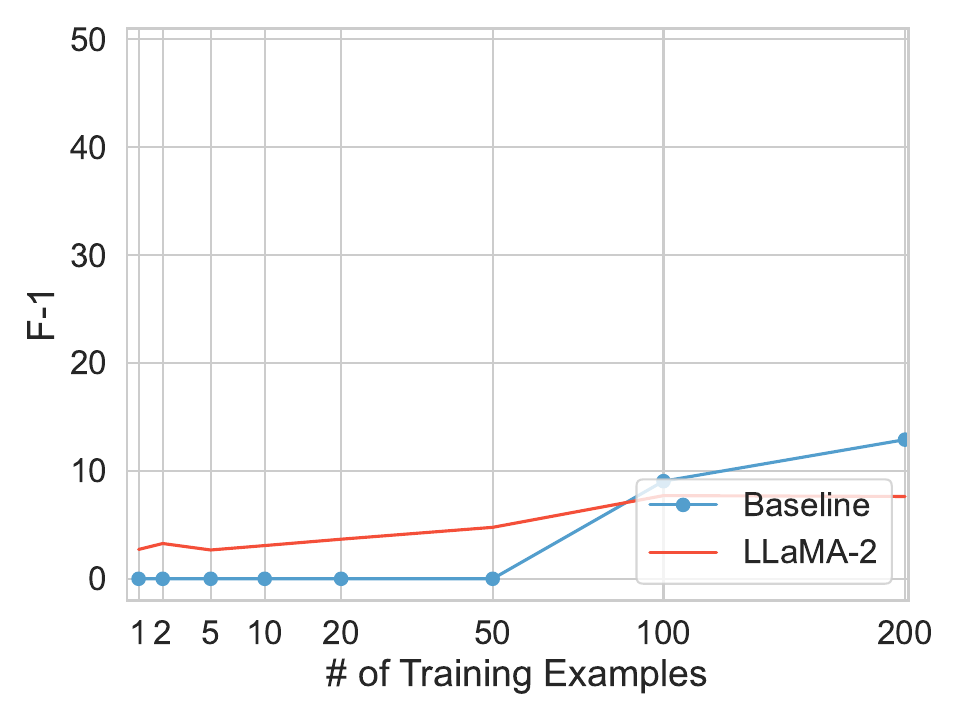} \\
        \textbf{(a)} Coreference & \textbf{(b)} Temporal & \textbf{(c)} Causal \\[3pt]
    \end{tabular}
    \begin{tabular}{cccc}
        \includegraphics[width=0.34\textwidth]{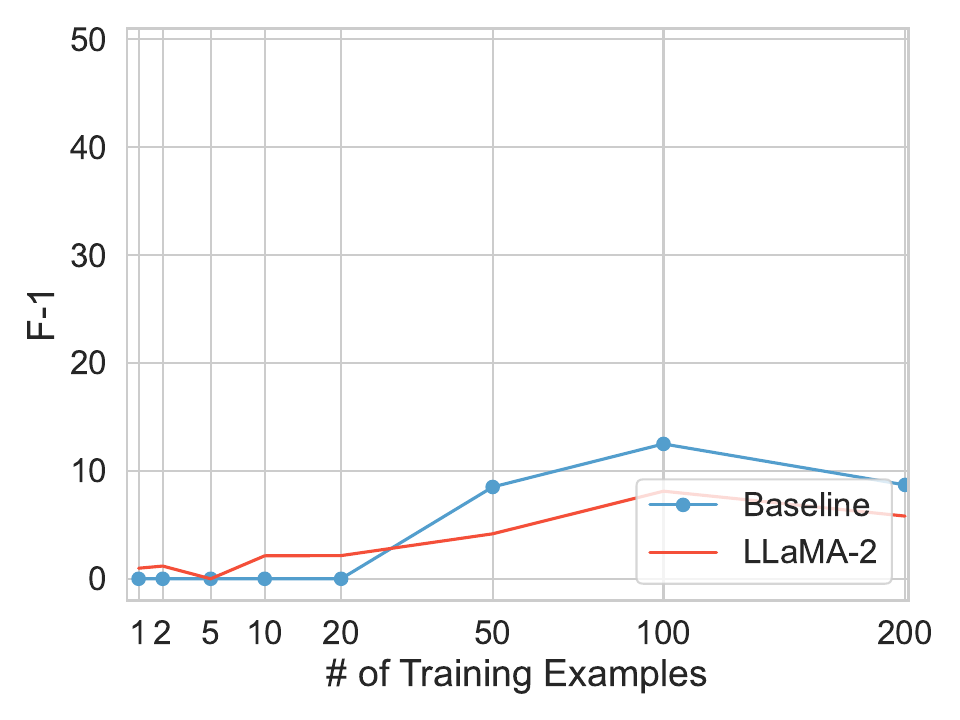} &
        \includegraphics[width=0.34\textwidth]{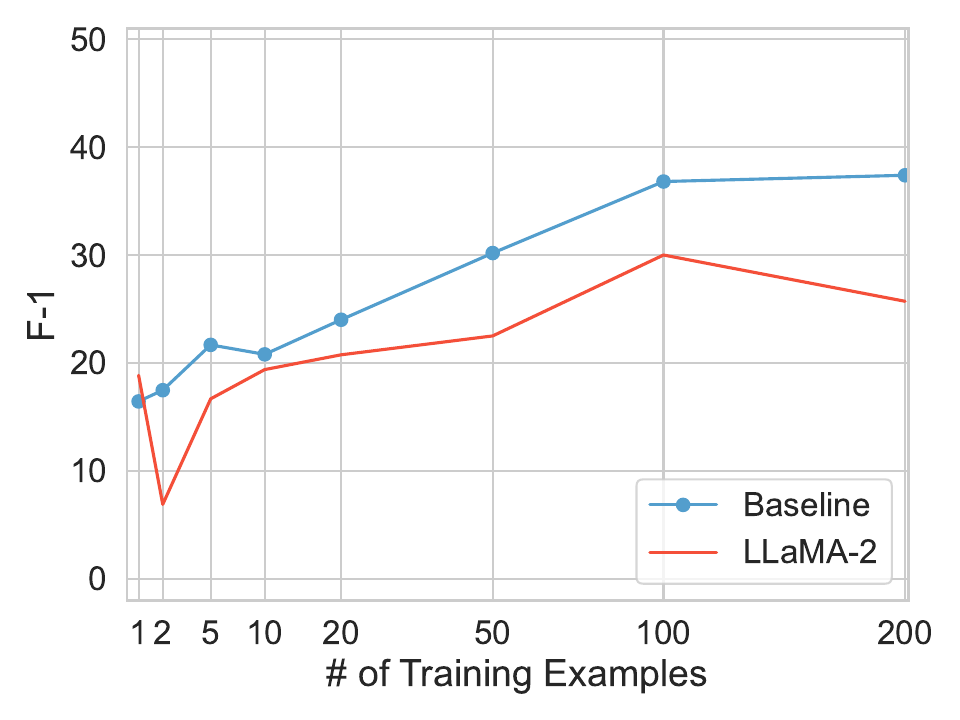} \\
        \textbf{(d)} Subevent & \textbf{(e)} Overall \\[3pt]
    \end{tabular}
    \vspace{-0.2cm}
    \caption{F-1 scores for coreference, temporal, causal, subevent relations, and overall performance. The blue lines represent the baseline's performance trained with different numbers of documents. The red lines represent the performance of LLaMA-2 fine-tuned with different numbers of documents.}
    \label{fig:f1-curves}}
\end{figure*}

\vspace{-0.1in}

\paragraph{SFT with a Varying Size of Training Data} 
Will LLMs perform better when we fine-tune them with training data? How will LLMs compare to the smaller baseline model when fine-tuned using the same amount of data?
We answer this question by varying the amount of training data for the supervised baseline method and meanwhile fine-tuning LLaMA-2
with the same amount of training data. In this experiment, LLaMA-2 is fine-tuned in the pairwise format as described in Section~\ref{sec:pairwise} to match with the baseline model \cite{wang-etal-2022-maven} which treats the ERE tasks as pairwise classification tasks. Detailed hyperparameters for fine-tuning LLaMA-2 can be found in Appendix~\ref{sec:hyperparameter}.

Figure~\ref{fig:f1-curves} shows the result comparisons between LLaMA-2 and the smaller baseline model. First, SFT certainly improves the performance of LLaMA-2. However, it still underperforms the smaller baseline method for all the relation types as the number of used training documents increases. 
LLaMA-2 typically requires twice more training data to reach the same overall performance as the smaller baseline model. 
LLaMA-2 only outperforms the smaller baseline method when the available training data is very limited, as LLaMA-2 benefits from its zero-shot capability emergent from large-scale pre-training. It also deserves mentioning that fine-tuning LLMs is much more expensive than fine-tuning the smaller baseline language model \textit{RoBERTa}. For example, fine-tuning using 200 training documents for 3 epochs requires approximately 72 hours for LLaMA-2 7B, while only about one hour is needed for the \textit{roberta-base} baseline model. 

\section{Discussion}

\subsection{GPT-3.5 and LLaMA-2 Struggle to Follow the Prompt Consistently}
During the test of GPT-3.5 and LLaMA-2, we notice that both models  create events or event relations that do not exist in text. 
 
In addition, both models occasionally have difficulties in generating formatted answers, and the outputs may consist of random words from the document rather than Event or TIMEX ID or even event trigger words.

As GPT-3.5 has achieved overall better performance compared to LLaMA-2, we conduct further analysis on model performance mainly based on the predictions of GPT-3.5 on the 10 validation documents.

\subsection{GPT-3.5 failed to learn transitivity rules} 
\begin{table}[htb]
\small
\footnotesize
\scalebox{1.0}{\begin{tabular}{l | c c c c}
    \midrule
     & &FP & FN & \begin{tabular}[c]{@{}c@{}}Transitivity\\Fixable\end{tabular} \\
     \midrule
     \multirow{3}{*}{whole doc} & Temporal & $21.63$ & $66.33$ & $7.40$ \\
     &Causal & $64.33$ & $32.48$ & $0$ \\
     &Subevent & $77.35$ & $20.99$ & $-$ \\
     \midrule
     \multirow{3}{*}{10-shot} & Temporal & $23.85$ & $63.42$ & $6.02$ \\
     &Causal & $59.71$ & $34.29$ & $0.43$ \\
     &Subevent & $82.77$ & $15.55$ & $-$ \\
     \midrule
\end{tabular}}
\vspace{-0.1in}
\caption{Rates (\%) of various errors in GPT-3.5 predictions with Whole doc and 10-shot prompt settings. False positives (FP) and false negatives (FN) are indicated. Transitivity fixable denotes the enhancement in F-1 score achieved by incorporating transitive relations through true positives.}
\vspace{-0.2in}
\label{tab:transitivity_error_analysis_table}
\end{table}
By manually examine the output of GPT-3.5 on the 10 validation documents, we notice that this model failed to learn the transitivity rules from the provided examples. When the output from GPT-3.5 contains tuples like (Event\_0, BEFORE, Event\_1) and (Event\_1, BEFORE, Event\_2), the tuple (Event\_0, BEFORE, Event\_2) can be inferred from the existing predictions. However, GPT-3.5 failed to include such tuples that can be inferred using transitivity rules. On the contrary, GPT-3.5 sometimes predict the opposite of the correct tuple. In this case, instead of (Event\_0, BEFORE, Event\_2), (Event\_2, BEFORE, Event\_0) will be predicted by GPT model, which clearly violates  transitivity rules. 

According to \citet{wang-etal-2022-maven}, 88.8\% temporal relations and 23.9\% causal relations can be inferred with transitivity rules in the \mavenere dataset. Failure to follow the transitivity rules detriments GPT's performance. 
Moreover, Table \ref{tab:transitivity_error_analysis_table} highlights that false negatives and false positives emerge as the predominant error types, indicating that GPT-3.5 faces challenges in accurately discerning the presence or absence of relationships. Notably, in the context of temporal relationships, a noteworthy increase in the F1 score is observed when incorporating all transitivity-inferred relations. This implies that GPT-3.5, indeed falls short of capturing a comprehensive array of transitive relations.

\subsection{Event Pairs with a Varying Distance}

\vspace{-0.15in}
\begin{table}[htb]
\small
\footnotesize
\begin{center}
\scalebox{0.75}{\begin{tabular}{c | c c |c c c c c}
    \midrule
     & \begin{tabular}[c]{@{}c@{}}Intra\\($<1$)\end{tabular}& \begin{tabular}[c]{@{}c@{}}Inter\\($>=1$)\end{tabular} & $1$ & $2$ & $3$ & $4$ & $>=5$ \\
     \midrule
     Temporal & $25.45$ & $12.39$ & $19.97$ & $17.70$ & $10.23$ & $11.76$ & $6.87$ \\
     Causal & $8.92$ & $7.02$ & $11.02$ & $7.69$ & $0$ & $0$ & $8.70$\\
     Subevent & $4.65$ & $2.56$ & $5.26$ & $0$ & $0$ & $0$ & $0$ \\
     \midrule
\end{tabular}}
\end{center}
\vspace{-0.1in}
\caption{GPT-3.5 performance (F-1 score) using Iterative Prediction 10-shot prompt on data groups with varying distances (measured in \#sentences) between related events.}
\vspace{-0.1in}
\label{tab:distance_table}
\end{table}
We investigate model performance on predicting intra- and inter-sentence event relations separately. 
Given that event coreference resolution relies on undividable 
clusters, we mainly analyze performance on the other three tasks.
As shown in Table~\ref{tab:distance_table}, GPT-3.5 is more capable to capture the relations between events that appear in the same sentence (intra-sentence) and otherwise struggles with capturing the inter-sentence event relations. 

We also investigate the impact of sentence distance on model performance for inter-sentence cases. 
As shown in \ref{tab:distance_table}, the performance on temporal relation extraction decreases quickly as the number of sentences between two events increases; the performance on causal relation extraction also decreases a little when the number of sentences in between increases from one to two, but then the performance seems to remain stable when we further consider causal relations with five or more sentences in between. 
Overall, we observe lower performance on event pairs with greater distances.

\subsection{Event Pairs in Contexts of Varying Event Densities}

\begin{table}[htb]
\small
\footnotesize
\begin{center}
\scalebox{0.8}{\begin{tabular}{l | c c c c}
    \midrule
     & & $2$ & $3$ & $>=4$ \\
     \midrule
     \multirow{3}{*}{10-shot} & Temporal & $31.25$ & $25.24$ & $24.55$ \\
     &Causal & $17.39$ & $9.52$ & $6.52$ \\
     &Subevent & $0$ & $16.0$ & $0$ \\
     \midrule
\end{tabular}}
\end{center}
\vspace{-0.1in}
\caption{GPT-3.5 performance (F-1 score) using the Iterative Prediction 10-shot prompt on data groups with different event densities (measured by \# of events within the same sentence).} 
\vspace{-0.1in}
\label{tab:density_error_analysis_sentence_table}
\end{table}
We investigate the impact of event density on model performance by examining the predictions of GPT-3.5 on the 10 validation documents. 
We only consider event pairs within the same sentence. 
Event density is measured as the number of event and TIMEX mentions appeared in one sentence. As shown in Table~\ref{tab:density_error_analysis_sentence_table}, the performance of GPT-3.5 on temporal and causal relations decreases quickly as the event density increases, indicating GPT-3.5 struggles to capture event relations when the complexity of the context increases.

\section{Conclusion}
In this study, we systematically evaluated the effectiveness of LLMs in performing discourse-level ERE tasks featuring lengthy documents and intricate relations. Our experiments using multiple prompt patterns uncover a noteworthy underperformance of LLMs when compared to the baseline established through supervised learning. 
Even with supervised fine-tuning, LLMs like LLaMA-2 still underperform the much smaller supervised baseline model when trained on the same amount of data.
Furthermore, our quantitative and qualitative analyses revealed that LLMs face challenges in obeying transitivity rules, 
capturing inter-sentence relations or relations with a long distance, as well as comprehending context with dense event mentions. 
For future work, we will further investigate these challenges and develop methods for enabling LLMs to better address some of these issues in event relation extraction. 

\section*{Limitation}
Although we tried several different prompt patterns, there is still a chance that there exists better prompt to be used to assist GPT to solve the ERE task better. Meanwhile, OpenAI constantly update the GPT models that can be accessed throught the API, making it hard to reproduce the results if older models are deprecated. While OpenAI has offered preliminary introductions to various versions of GPT models, the specific implementation details remain obscure. This opacity hampers thorough analysis of why distinct versions of GPT models exhibit varying performance levels and how each data set and training technique influences models' performance. 
Finally, OpenAI API is a paid service, conducting experiments can get expensive depending on the task and design of the evaluation, making it not accessible to larger community. We are also limited by the cost and response time of OpenAI API.

For LLaMA-2 models, larger models (13B and 70B) may have better performance, but we leave the thorough study of LLaMA-2 models for potential future works.

\section*{Ethics and Broader Impact}
We are aware that such study is very expensive and not very accessible to some researchers in the NLP community as OpenAI API is a paid service and is restricted in many countries. Not all researchers in our community can afford thousands of dollars or even more to run such experiments. Experiments on exclusive models or API may further detriment the inclusiveness of NLP community. Therefore, we hope our work can provide insights to readers with limited resources and inspire them in other works. However, by no means that we are advocating the NLP community to include closed-source LLMs as the baseline for any of the future work as studying the performance and behavior of closed-source models can be extremely difficult. We aim for our work to serve as a valuable resource for readers, helping them make decisions as they leverage LLMs for complex ERE tasks at discourse-level.

\section*{Acknowledgements}
We would like to thank the anonymous reviewers
for their valuable feedback and input. We gratefully acknowledge support from National Science Foundation via the award IIS-1942918. Portions of this research were conducted
with the advanced computing resources provided
by Texas A\&M High-Performance Research Computing.

\bibliography{anthology,custom}

\newpage
\section*{Appendix} \label{sec:appendix}
\addcontentsline{toc}{subsection}{Appendix}
\renewcommand{\thesubsection}{\Alph{subsection}}
\subsection{Hyperparameter and Compute Detail} \label{sec:hyperparameter}
We use the default hyperparameters for \textit{gpt-3.5-turbo-16k} and \textit{gpt-4-1106-preview} with the temperature set to 1, top\_p set to 1, frequency penalty set to 0, and presence penalty set to 0. It takes around 48 - 60 hours to run though the test set of \mavenere for inference with \textit{gpt-3.5-turbo-16k}. For LLaMA-2, we set temperature to 0.6, and top\_p to 0.9, following the llama-recipes\footnote{\url{https://github.com/meta-llama/llama-recipes/tree/main}} GitHub repository maintained by Meta. It takes around 36 - 48 hours to run though the test set of \mavenere for inference with \textit{Llama-2-7b-chat-hf}.

For the baseline model, we train the model following the hyperparameters in \citet{wang-etal-2022-maven}. We train the model with the learning rate sets to $1e-5$ for the RoBERTa model, the learning rate sets to $1e-5$ for the classification head, and the batch size sets to 4. For coreference, temporal, and causal resolutions, we train the model for 50 epochs. For subevent resolution, we train the model for 20 epochs. The training and inference time in total varies from 30 minutes to 2.5 hours depending on the relation type.

For SFT with \textit{Llama-2-7b-chat-hf}, we fine-tune the model for 3 epochs, with a learning rate of $2e-4$, weght decay of $0.001$ for AdamW optimizer. 4-bit quantization is used to save memory space, and LoRA with 64 attention dimension, and 0.1 dropout rate is used for speeding up the training process.

\subsection{Models' Versions}
\label{sec:model versions}
The \textit{gpt-3.5-turbo-16k} currently points to \textit{gpt-3.5-turbo-0613}, which is a snapshot of \textit{gpt-3.5-turbo} from June 13th 2023 and will be deprecated on June 13, 2024, according to the OpenAI website \url{https://platform.openai.com/docs/models}. All our final experiment runs were conducted in a relatively focused period of time, in November 2023, so the model we used is the \textit{gpt-3.5-turbo-0613}.

The \textit{gpt-4-1106-preview} was released on November 6th, 2023, and has knowledge of world events up to April 2023, according to the OpenAI website \url{https://platform.openai.com/docs/models}. All our final experiment runs were conducted in a relatively focused period of time, in January 2024.

The LLaMA-2 model used in this paer is \textit{Llama-2-7b-chat-hf}, which can be accessed through Huggingface \url{https://huggingface.co/meta-llama/Llama-2-7b-chat-hf}.

\subsection{GPT-3.5 Experiments Cost}
We use the \textit{gpt-3.5-turbo-16k} model from \href{https://platform.openai.com/docs/models}{OpenAI API}. The cost for \textit{gpt-3.5-turbo-16k} is \$0.001 per 1k tokens for input and \$0.002 per 1k tokens for output. We run through the \mavenere test set (857 documents) for 5 times since we run through the whole test set once for each of the whole doc, 1-shot, 2-shot, 5-shot, and 10-shot prompt. The total cost is around \$1650, resulting \$330 on average for each different prompt pattern used, \$0.385 on average for annotating a document.

\subsection{Prompt Patterns} \label{sec:prompts patterns}
Here, we provide some examples for the whole doc and $n$-shot prompt described in Sec~\ref{sec: prompting}. The system prompts for coreference, temporal, causal and subevent relations are shown in Table~\ref{table:causal and subevent system prompt table}. We show an example of whole doc prompt for coreference and temporal relations in Table~\ref{table:whole doc prompt table} , and an example of 2-shot prompt for coreference and temporal relations in Table~\ref{table:few shot prompt table}. Causal and subevent relations follow the same pattern as temporal relation. 
\begin{table}[h]
\centering
\scalebox{0.6}{
\begin{tabular}{m{1.8cm}|p{10cm}} 
    \toprule
    \centering Relation Type & \centering \textbf{System} \tabularnewline [0.5ex]
    \midrule
    \centering\arraybackslash Coreference & You are an annotator for the MAVEN-ERE dataset. Your task is to extract event coreference relations between event mentions from given documents, where all event and TIMEX mentions are given in [ ]. Imitate the given example to find coreference relations between event mentions. The last sentence of the context is not annotated. You should find all the relations among the new mentions in the last sentence with mentions in all previous sentences. Predict the relations in this format: Event\_1 COREFERENCE Event\_0; SHIFT; means moving on to the next sentence. Always add SHIFT; at the end of prediction. \tabularnewline
    \midrule
    Temporal & You are an annotator for the MAVEN-ERE dataset. Your task is to extract temporal relations between event mentions from given documents, where all event and TIMEX mentions are given in [ ]. There are 6 types of temporal relations: BEFORE, CONTAINS, OVERLAP, BEGINS-ON, ENDS-ON, and SIMULTANEOUS. Imitate the given example to find temporal relations between event and TIMEX mentions. The last sentence of the context is not annotated. You should find all the relations among the new mentions in the last sentence with mentions in all previous sentences. Predict the relations in this format: Event\_1 BEFORE Event\_0; SHIFT; means moving on to the next sentence. Always add SHIFT; at the end of prediction.\tabularnewline
    \midrule
    \centering Causal & You are an annotator for the MAVEN-ERE dataset. Your task is to extract causal relations between event mentions from given documents, where all event and TIMEX mentions are given in [ ]. There are 2 types of causal relations: CAUSE, and PRECONDITION. Imitate the given example to find causal relations between event and TIMEX mentions. The last sentence of the context is not annotated. You should find all the relations among the new mentions in the last sentence with mentions in all previous sentences. Predict the relations in this format: Event\_1 CAUSE Event\_0; SHIFT; means moving on to the next sentence. Always add SHIFT; at the end of prediction.\tabularnewline
    \midrule
    \centering Subevent & You are an annotator for the MAVEN-ERE dataset. Your task is to extract subevent relations between event mentions from given documents, where all event and TIMEX mentions are given in [ ]. Imitate the given example to find subevent relations between event and TIMEX mentions. The last sentence of the context is not annotated. You should find all the relations among the new mentions in the last sentence with mentions in all previous sentences. Predict the relations in this format: Event\_1 SUBEVENT Event\_0; SHIFT; means moving on to the next sentence. Always add SHIFT; at the end of prediction.\tabularnewline [0.5ex]
    \bottomrule
\end{tabular}}
\caption{System prompts for causal and subevent relations.}
\label{table:causal and subevent system prompt table}
\end{table}
\begin{table*}[h]
\centering
\scalebox{0.7}{
\begin{tabular}{m{1.1cm}|p{10cm}|p{10cm}} 
    \toprule
    \centering Relation type & \centering \textbf{Coreference} & \centering \textbf{Temporal} \tabularnewline [0.5ex]
    \midrule
    \centering Prompt & \includegraphics[scale=0.01]{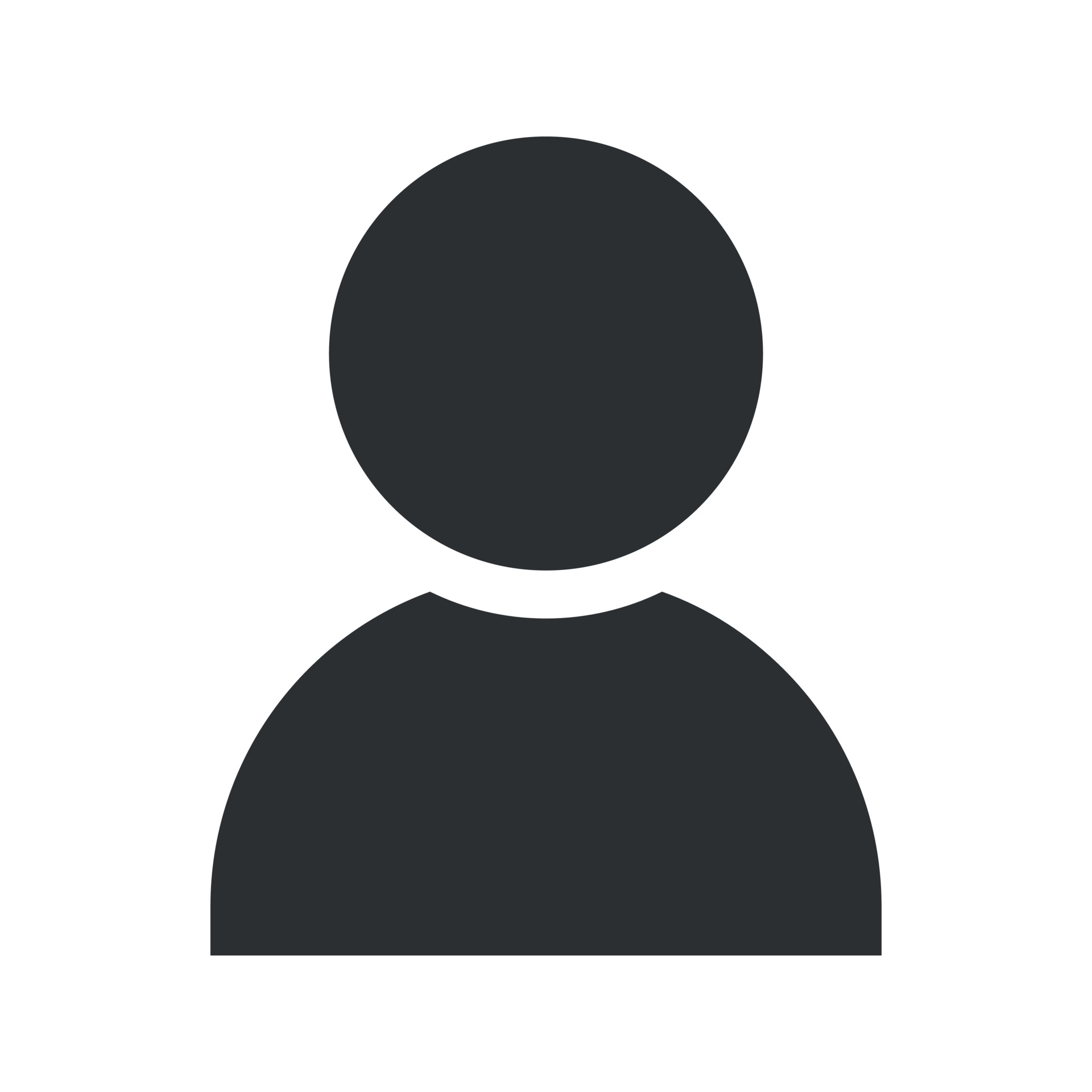} The [ 0 Expedition Event\_0 ] of the Thousand ( Italian `` Spedizione dei Mille '' ) was an event of the Italian Risorgimento that [ took place Event\_1 ] in [ 1860 TIMEX\_0 ]. a corps of volunteers led by giuseppe garibaldi [ sailed Event\_2 ] from quarto, near genoa ( now quarto dei mille ) and [ landed Event\_3 ] in marsala, sicily, in order to [ 1 conquer Event\_4 ] the kingdom of the two sicilies, [ ruled Event\_5 ] by the house of bourbon-two sicilies. The project was an ambitious and risky [ 0 venture Event\_6 ] [ aiming Event\_7 ] to [ 1 conquer Event\_8 ], with a thousand men, a kingdom with a larger regular army and a more powerful navy. \newline
    \includegraphics[scale=0.02]{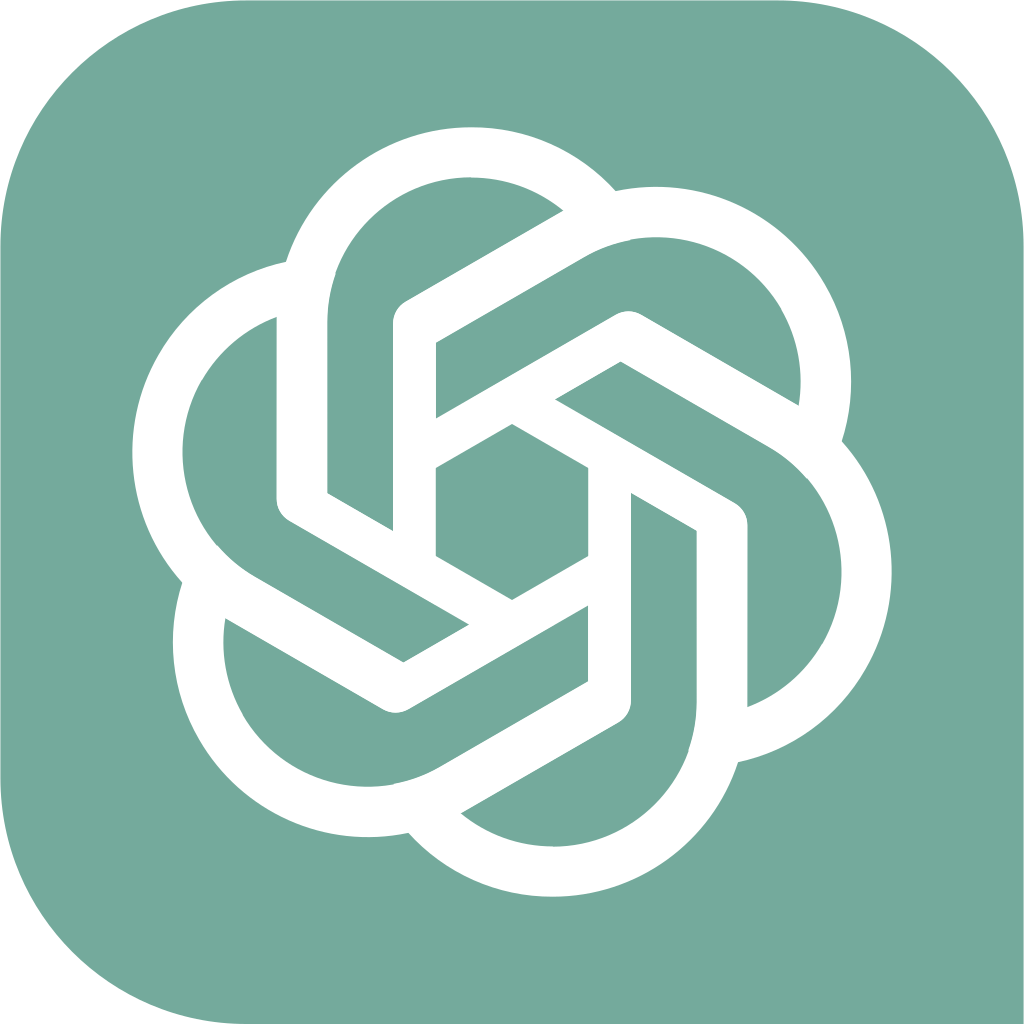}/\includegraphics[scale=0.06]{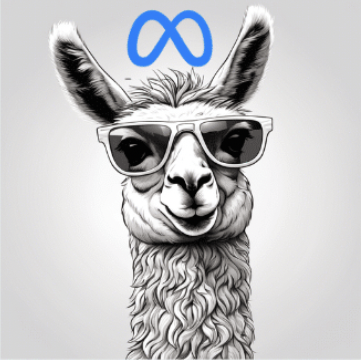} Event\_9 COREFERENCE 0; Event\_13 COREFERENCE 1; SHIFT; \newline
    \includegraphics[scale=0.01]{images/human_user_logo.jpg} The King David Hotel [ 0 bombing Event\_0 ] was a terrorist [ 0 attack Event\_1 ] [ carried out Event\_2 ] on [ Monday, July 22, 1946 TIMEX\_0 ], by the militant right-wing Zionist underground organization the Irgun on the British administrative headquarters for Palestine, which was housed in the southern wing of the King David Hotel in Jerusalem during the Jewish insurgency in Mandatory Palestine. 91 people of various nationalities were [ killed Event\_3 ], and 46 were [ injured Event\_4 ]. the hotel was the site of the central offices of the british mandatory authorities of palestine, principally the secretariat of the government of palestine and the headquarters of the british armed forces in palestine and transjordan. \textcolor{red}{When [ planned Event\_5 ], the [ attack Event\_6 ] had the [ approval Event\_7 ] of the Haganah, the principal Jewish paramilitary group in Palestine, though, unbeknownst to the Irgun, this had been [ cancelled Event\_8 ] by the time the [ operation Event\_9 ] was [ carried out Event\_10 ].} &

    \includegraphics[scale=0.01]{images/human_user_logo.jpg} The [ Expedition Event\_0 TIMEX\_0 CONTAINS Event\_0;Event\_5 OVERLAP Event\_0; ] of the Thousand ( Italian `Spedizione dei Mille' ) was an event of the Italian Risorgimento that [ took place Event\_1 TIMEX\_0 CONTAINS Event\_1;Event\_5 OVERLAP Event\_1;Event\_0 SIMULTANEOUS Event\_1; ] in [ 1860 TIMEX\_0 Event\_5 OVERLAP TIMEX\_0; ]. A corps of volunteers led by giuseppe garibaldi [ sailed Event\_2 ] from quarto, near genoa ( now quarto dei mille ) and [ landed Event\_3 ] in marsala, sicily, in order to [ conquer Event\_4 ] the kingdom of the two sicilies, [ ruled Event\_5 ] by the house of bourbon-two sicilies. \newline 
    \includegraphics[scale=0.02]{images/gpt_logo.png}/\includegraphics[scale=0.06]{images/llama2_logo.png} Event\_0 CONTAINS Event\_2; Event\_1 CONTAINS Event\_2; TIMEX\_0 CONTAINS Event\_2; Event\_0 CONTAINS Event\_3; Event\_1 CONTAINS Event\_3; ... Event\_5 CONTAINS Event\_3; SHIFT;\newline
    \includegraphics[scale=0.01]{images/human_user_logo.jpg} The Cherry Valley massacre was an attack by British and Iroquois forces on a fort and the village of Cherry Valley in eastern New York on [ November 11, 1778 TIMEX\_0 TIMEX\_1 CONTAINS TIMEX\_0; ], during [ the American Revolutionary War TIMEX\_1  ]. \textcolor{red}{It has been [ described Event\_0 ] as one of the most horrific frontier massacres of the war.} \tabularnewline [0.5ex]
    \bottomrule
\end{tabular}}
\caption{An example of the whole doc prompt on coreference and temporal relations. Causal and subevent relations follow the same pattern as temporal relation. The sentence in red is the sentence being queried.}
\label{table:whole doc prompt table}
\end{table*}
\begin{table*}[h]
\centering
\scalebox{0.7}{
\begin{tabular}{m{1cm}|p{10cm}|p{10cm}} 
    \toprule
    \centering Relation type & \centering \textbf{Coreference} & \centering \textbf{Temporal} \tabularnewline [0.5ex]
    \midrule
    \centering Prompt & \includegraphics[scale=0.01]{images/human_user_logo.jpg} The Cherry Valley massacre was an attack by British and Iroquois forces on a fort and the village of Cherry Valley in eastern New York on [ November 11, 1778 TIMEX\_0 ], during [ the American Revolutionary War TIMEX\_1 ].\newline
    \includegraphics[scale=0.02]{images/gpt_logo.png}/\includegraphics[scale=0.06]{images/llama2_logo.png} SHIFT; \newline
    \includegraphics[scale=0.01]{images/human_user_logo.jpg} The Cherry Valley massacre was an attack by British and Iroquois forces on a fort and the village of Cherry Valley in eastern New York on [ November 11, 1778 TIMEX\_0 ], during [ the American Revolutionary War TIMEX\_1 ]. It has been [ described Event\_0 ] as one of the most horrific frontier massacres of the war. \newline
    \includegraphics[scale=0.02]{images/gpt_logo.png}/\includegraphics[scale=0.06]{images/llama2_logo.png} SHIFT; \newline
    
    \includegraphics[scale=0.01]{images/human_user_logo.jpg} The King David Hotel [ bombing Event\_0 ] was a terrorist [ attack Event\_1 ] [ carried out Event\_2 ] on [ Monday, July 22, 1946 TIMEX\_0 ], by the militant right-wing Zionist underground organization the Irgun on the British administrative headquarters for Palestine, which was housed in the southern wing of the King David Hotel in Jerusalem during the Jewish insurgency in Mandatory Palestine. \newline
    \includegraphics[scale=0.02]{images/gpt_logo.png}/\includegraphics[scale=0.06]{images/llama2_logo.png} Event\_1 COREFERENCE Event\_0; SHIFT; \newline
    \includegraphics[scale=0.01]{images/human_user_logo.jpg} The King David Hotel [ 0 bombing Event\_0 ] was a terrorist [ 0 attack Event\_1 ] [ carried out Event\_2 ] on [ Monday, July 22, 1946 TIMEX\_0 ], by the militant right-wing Zionist underground organization the Irgun on the British administrative headquarters for Palestine, which was housed in the southern wing of the King David Hotel in Jerusalem during the Jewish insurgency in Mandatory Palestine. 91 people of various nationalities were [ killed Event\_3 ], and 46 were [ injured Event\_4 ].\newline
    \includegraphics[scale=0.02]{images/gpt_logo.png}/\includegraphics[scale=0.06]{images/llama2_logo.png} SHIFT; \newline
    
    \includegraphics[scale=0.01]{images/human_user_logo.jpg} The [ Battle Event\_0 ] of Orthez ( [ 27 February 1814 TIMEX\_0 ] ) saw the Anglo-Portuguese Army under Field Marshal Arthur Wellesley, Marquess of Wellington [ attack Event\_1 ] an Imperial French army [ led Event\_2 ] by Marshal Nicolas Soult in southern France. \textcolor{red}{The outnumbered French [ repelled Event\_3 ] several Allied [ assaults Event\_4 ] on their right flank, but their center and left flank were [ overcome Event\_5 ] and Soult was compelled to [ retreat Event\_6 ].}&

    \includegraphics[scale=0.01]{images/human_user_logo.jpg} The Cherry Valley massacre was an attack by British and Iroquois forces on a fort and the village of Cherry Valley in eastern New York on [ November 11, 1778 TIMEX\_0 ], during [ the American Revolutionary War TIMEX\_1 ].\newline 
    \includegraphics[scale=0.02]{images/gpt_logo.png}/\includegraphics[scale=0.06]{images/llama2_logo.png} TIMEX\_1 CONTAINS TIMEX\_0; SHIFT;\newline
    \includegraphics[scale=0.01]{images/human_user_logo.jpg} The Cherry Valley massacre was an attack by British and Iroquois forces on a fort and the village of Cherry Valley in eastern New York on [ November 11, 1778 TIMEX\_0 TIMEX\_1 CONTAINS TIMEX\_0; ], during [ the American Revolutionary War TIMEX\_1  ]. It has been [ described Event\_0 ] as one of the most horrific frontier massacres of the war. \newline 
    \includegraphics[scale=0.02]{images/gpt_logo.png}/\includegraphics[scale=0.06]{images/llama2_logo.png}SHIFT;\newline

    \includegraphics[scale=0.01]{images/human_user_logo.jpg} The United States occupation of Nicaragua from [ 1912 TIMEX\_0 ] to [ 1933 TIMEX\_1 ] was part of the Banana Wars, when the US military intervened in various Latin American countries from [ 1898 TIMEX\_2 ] to [ 1934 TIMEX\_3 ]. \newline 
    \includegraphics[scale=0.02]{images/gpt_logo.png}/\includegraphics[scale=0.06]{images/llama2_logo.png} TIMEX\_0 BEFORE TIMEX\_1; TIMEX\_2 BEFORE TIMEX\_0; TIMEX\_2 BEFORE TIMEX\_1; TIMEX\_0 BEFORE TIMEX\_3; TIMEX\_1 BEFORE TIMEX\_3; TIMEX\_2 BEFORE TIMEX\_3; SHIFT;\newline
    \includegraphics[scale=0.01]{images/human_user_logo.jpg} The United States occupation of Nicaragua from [ 1912 TIMEX\_0 TIMEX\_2 BEFORE TIMEX\_0; ] to [ 1933 TIMEX\_1 TIMEX\_0 BEFORE TIMEX\_1;TIMEX\_2 BEFORE TIMEX\_1;Event\_0 BEFORE TIMEX\_1;TIMEX\_4 BEFORE TIMEX\_1; ] was part of the Banana Wars, when the US military intervened in various Latin American countries from [ 1898 TIMEX\_2  ] to [ 1934 TIMEX\_3 TIMEX\_0 BEFORE TIMEX\_3;TIMEX\_1 BEFORE TIMEX\_3;TIMEX\_2 BEFORE TIMEX\_3;Event\_0 BEFORE TIMEX\_3;TIMEX\_4 BEFORE TIMEX\_3; ]. The formal occupation [ began Event\_0 ] in [ 1912 TIMEX\_4 ], even though there were various other assaults by the U.S. in Nicaragua throughout this period.\newline 
    \includegraphics[scale=0.02]{images/gpt_logo.png}/\includegraphics[scale=0.06]{images/llama2_logo.png}TIMEX\_0 BEFORE Event\_0; Event\_0 BEFORE TIMEX\_1; TIMEX\_2 BEFORE Event\_0; Event\_0 BEFORE TIMEX\_3; TIMEX\_0 BEFORE TIMEX\_4; TIMEX\_4 BEFORE TIMEX\_1; TIMEX\_2 BEFORE TIMEX\_4; TIMEX\_4 BEFORE TIMEX\_3; TIMEX\_4 CONTAINS Event\_0; SHIFT;\newline
    
    \includegraphics[scale=0.01]{images/human_user_logo.jpg} The [ Battle Event\_0 TIMEX\_0 SIMULTANEOUS Event\_0; ] of Malacca ( [ 2 August 1640 – 14 January 1641 TIMEX\_0  ] ) was a successful attempt by the Dutch to [ capture Event\_1 Event\_0 BEFORE Event\_1;TIMEX\_0 BEFORE Event\_1;] Malacca from the Portuguese. \textcolor{red}{In [ the early 17th century TIMEX\_1 ], the Dutch East India Company (Verenigde Oostindische Compagnie) [ began Event\_2 ] the campaign to [ destroy Event\_3 ] Portuguese power in the East.}
 \tabularnewline [0.5ex]
    \bottomrule
\end{tabular}}
\caption{An example of the 2-shot prompt on coreference and temporal relations. Causal and subevent relations follow the same pattern as temporal relation. The sentence in red is the sentence being queried.}
\label{table:few shot prompt table}
\end{table*}

We also show some examples for the two other prompt patterns: Bulk Prediction and Event Ranking. Examples for these two prompt patterns are shown in Table~\ref{table:prompt table}. We don't choose to use the Bulk Prediction prompt is because the performance is overall the worst comparing to other prompts when testing on the first 10 documents of the validation set. Event Ranking has relatively good performance on the first 10 validation document but require much more number of queries. We estimate a total cost of \$3,200 if use the event ranking prompt to run through the whole test set once. Event Ranking prompt is also very time-consuming with an estimated 400 hours to run through the test set. Since using event ranking prompt is both financially and time-wise impossible, we don't choose to use it. 
\begin{table*}[h]
\centering
\scalebox{0.7}{
\begin{tabular}{m{1cm}|p{10cm}|p{10cm}} 
    \toprule
    \centering Prompt Method & \centering \textbf{Bulk Prediction} & \centering \textbf{Event Ranking} \tabularnewline [0.5ex]
    \midrule
    \centering\arraybackslash System & You are an annotator for the MAVEN-ERE dataset. Your task is to extract event coreference, temporal, causal, and subevent relations between event and TIMEX mentions from given documents, where all event and TIMEX mentions are given. Coreference and subevent relations are binary. For temporal relations, there are 6 types: ... For causal relations, there are 2 types: ... Note that the order of the events matter. SIMULTANEOUS and BEGINS-ON are bidirectional relations. If there is no relations, return an empty array. You should always  finish the answer instead of using '...'. & 
    You are an annotator for the MAVEN-ERE dataset. Your task is to extract event coreference, temporal, causal, and subevent relations between event and TIMEX mentions from given documents, where all event and TIMEX mentions are given in [] after triggering words. All predictions should be an array with elements being EVENT and TIMEX mentions given in [] from document.\tabularnewline [0.5ex]
    \midrule
    \centering Prompt & \includegraphics[scale=0.01]{images/human_user_logo.jpg} This is the document: The Expedition \text{[Event\_0]} of the Thousand (Italian `Spedizione dei Mille') was ... in 1860 \text{[TIMEX\_0]}... distribution \text{[Event\_26]} and the \text{[Event\_27]} end of oppression. \newline 
    \includegraphics[scale=0.01]{images/human_user_logo.jpg} What are the temporal relations? \newline 
    \includegraphics[scale=0.02]{images/gpt_logo.png}/\includegraphics[scale=0.06]{images/llama2_logo.png}\text{\{BEFORE: [[Event\_17, Event\_0], [Event\_19, Event\_0]...],} \newline \text{CONTAINS: [[Event\_7, Event\_0], [TIMEX\_0, Event\_0]...],} \newline \text{OVERLAP: [[Event\_5, Event\_0], [Event\_5, Event\_1]...], ...} \newline \text{SIMULTANEOUS: [[Event\_0, Event\_1], [Event\_15, Event\_1]...]\}} \newline
    \includegraphics[scale=0.01]{images/human_user_logo.jpg} This is the document: The Cherry Valley massacre was an attack ... on November 11, 1778 \text{[TIMEX\_0]}, during ... leading to the 1779 \text{[TIMEX\_3]} Sullivan Expedition which drove \text{[Event\_14]} the Iroquois out of western New York. \newline 
    \includegraphics[scale=0.01]{images/human_user_logo.jpg} What are the temporal relations?&

    \includegraphics[scale=0.01]{images/human_user_logo.jpg} This is the document: The Expedition \text{[Event\_0]} of the Thousand (Italian `Spedizione dei Mille') was ... in 1860 \text{[TIMEX\_0]}... distribution \text{[Event\_26]} and the \text{[Event\_27]} end of oppression. \newline 
    \includegraphics[scale=0.01]{images/human_user_logo.jpg} List all mentions happened BEFORE \text{[Event\_0]}. If there is no relations, return an empty array. \newline 
    \includegraphics[scale=0.02]{images/gpt_logo.png}/\includegraphics[scale=0.06]{images/llama2_logo.png} \text{[Event\_17, Event\_19, Event\_20, Event\_21 ]} \newline
    \includegraphics[scale=0.01]{images/human_user_logo.jpg} This is the document: The Cherry Valley massacre was an attack ... on November 11, 1778 \text{[TIMEX\_0]}, during ... leading to the 1779 \text{[TIMEX\_3]} Sullivan Expedition which drove \text{[Event\_14]} the Iroquois out of western New York. \newline 
    \includegraphics[scale=0.01]{images/human_user_logo.jpg} List all mentions happened BEFORE \text{[Event\_0]}. If there is no relations, return an empty array. \tabularnewline [0.5ex]
    \bottomrule
\end{tabular}}
\caption{Other prompt patterns that are tried.}
\label{table:prompt table}
\end{table*}

\subsection{Detailed report of GPT-3.5, GPT-4 and LLaMA-2 Performance Over Validation Set}
\label{sec: detailed valid tables}
Here, we report the performance (precision, recall, and F-1 scores) of GPT-3.5 on the first 10 validation documents using the Iterative Prediction prompt patterns in Table~\ref{tab:valid-table}. We report the performance (precision, recall, and F-1 scores) of GPT-4 on the first 10 validation documents using the Iterative Prediction prompt patterns in Table~\ref{tab:gpt4-valid-table}. we report the performance (precision, recall, and F-1 scores) of LLaMA-2 on the first 10 validation documents using the Iterative Prediction prompt patterns in Table~\ref{tab:llama-valid-table}.
\begin{table*}[t!]
\small
\begin{center}
\scalebox{0.87}{\begin{tabular}{ lrrr|rrr|rrr|rrr} 
 \toprule
 & \multicolumn{3}{c|}{\textbf{MUC}} & \multicolumn{3}{c|}{\textbf{B$^{3}$}} & \multicolumn{3}{c|}{\textbf{CEAF$_{e}$}} & \multicolumn{3}{c}{\textbf{BLANC}} \\
 
\textbf{\textbf{Prompt}} & Precision & Recall & F-1 & Precision & Recall & F-1 & Precision & Recall & F-1 & Precision & Recall & F-1\\

 \midrule
 Bulk Prediction & & & & & & & & & & & &\\
 \hspace{.1in}\begin{tabular}[c]{@{}c@{}}whole doc\\(1-shot)\end{tabular} & $3.3$ & $14.3$ & $5.3$ & $77.7$ & $95.1$ & $85.5$ & $87.4$ & $71.0$ & $78.3$ & $51.4$ & $53.6$ & $51.9$\\
 \midrule
 Iterative Prediction & & & & & & & & & & & &\\
 \hspace{.1in}whole doc & $9.5$ & $28.6$ & $14.3$ & $85.3$ & $95.8$ & $90.2$ & $92.8$ & $82.4$ & $87.3$ & $52.6$ & $59.8$ & $53.8$\\ 
 \hspace{.1in}1-shot & $2.4$ & $7.1$ & $3.6$ & $84.6$ & $94.9$ & $89.4$ & $91.9$ & $82.0$ & $86.6$ & $50.2$ & $50.9$ & $50.0$\\ 
 \hspace{.1in}2-shot & $\mathbf{10.6}$ & $\mathbf{35.7}$ & $\mathbf{16.4}$ & $83.2$ & $\mathbf{96.0}$ & $89.2$ & $93.0$ & $80.8$ & $86.5$ & $52.3$ & $61.6$ & $53.4$\\ 
 \hspace{.1in}5-shot & $7.7$ & $21.4$ & $11.3$ & $\mathbf{86.0}$ & $95.6$ & $\mathbf{90.5}$ & $92.5$ & $83.2$ & $87.6$ & $\mathbf{52.9}$ & $\mathbf{61.9}$ & $\mathbf{54.3}$\\ 
 \hspace{.1in}10-shot & $5.1$ & $14.3$ & $7.6$ & $85.8$ & $95.3$ & $90.3$ & $\mathbf{93.2}$ & $\mathbf{83.9}$ & $\mathbf{88.3}$ & $50.9$ & $53.3$ & $51.2$\\
 \midrule
 Event Ranking & & & & & & & & & & & &\\
 \hspace{.1in}\begin{tabular}[c]{@{}c@{}}whole doc\\(1-shot)\end{tabular} & $4.4$ & $14.3$ & $6.7$ & $83.3$ & $95.1$ & $88.8$ & $88.9$ & $77.5$ & $82.8$ & $51.9$ & $53.8$ & $52.5$\\
\bottomrule

 \toprule
 & \multicolumn{3}{c|}{\textbf{Temporal}} & \multicolumn{3}{c|}{\textbf{Causal}} & \multicolumn{3}{c|}{\textbf{Subevent}} & \multicolumn{3}{c}{\textbf{Overall}} \\
 
 \textbf{\textbf{Prompt}} & Precision & Recall & F-1 & Precision & Recall & F-1 & Precision & Recall & F-1 & Precision & Recall & F-1\\
 \midrule
 Bulk Prediction & & & & & & & & & & & &\\
 \hspace{.1in}\begin{tabular}[c]{@{}c@{}}whole doc\\(1-shot)\end{tabular} & $7.2$ & $2.3$ & $3.4$ & $2.8$ & $2.4$ & $2.6$ & $3.0$ & $\mathbf{9.8}$ & $4.6$ & $17.0$ & $18.3$ & $16.5$\\
 \midrule
  Iterative Prediction & & & & & & & & & & & &\\
 \hspace{.1in}whole doc & $19.9$ & $8.6$ & $12.0$ & $2.2$ & $4.2$ & $2.9$ & $2.1$ & $7.3$ & $3.3$ & $21.1$ & $21.7$ & $19.9$\\
 \hspace{.1in}1-shot & $18.0$ & $10.4$ & $13.2$ & $5.0$ & $7.7$ & $6.1$ & $2.8$ & $\mathbf{9.8}$ & $4.3$ & $20.8$ & $21.7$ & $20.3$\\
 \hspace{.1in}2-shot & $18.5$ & $7.1$ & $10.2$ & $4.6$ & $6.0$ & $5.2$ & $3.2$ & $7.3$ & $4.4$ & $21.5$ & $22.2$ & $20.3$\\
 \hspace{.1in}5-shot & $16.3$ & $8.0$ & $10.8$ & $3.9$ & $4.8$ & $4.3$ & $0.0$ & $0.0$ & $0.0$ & $20.0$ & $19.6$ & $19.0$\\
 \hspace{.1in}10-shot & $\mathbf{22.1}$ & $10.6$ & $14.3$ & $\mathbf{6.2}$ & $\mathbf{10.1}$ & $\mathbf{7.7}$ & $2.0$ & $\mathbf{9.8}$ & $3.3$ & $\mathbf{22.3}$ & $23.1$ & $21.2$\\ 
 \midrule
Event Ranking & & & & & & & & & & & &\\
\hspace{.1in}\begin{tabular}[c]{@{}c@{}}whole doc\\(1-shot)\end{tabular} & $15.8$ & $\mathbf{34.1}$ & $\mathbf{21.6}$ & $4.8$ & $7.6$ & $5.9$ & $\mathbf{4.4}$ & $7.9$ & $\mathbf{5.7}$ & $20.5$ & $\mathbf{27.4}$ & $\mathbf{22.7}$\\
\bottomrule
\end{tabular}}
\end{center}
\vspace{-0.1in}
\caption{Event coreference, temproal, causal, and subevent resolution performances of \textit{gpt-3.5-turbo-16k} on \mavenere using different prompt patterns on the first 10 documents of the validation set. The numbers in \textbf{bold} indicate the best performance across different prompt patterns for \textit{gpt-3.5-turbo-16k}. Notice that the Event Ranking 1-shot prompt has the best overall performance and the Bulk Prediction 1-shot prompt has the worst overall performance.}
\label{tab:valid-table}
\end{table*}
\begin{table*}[t!]
\small
\begin{center}
\scalebox{0.87}{\begin{tabular}{ lrrr|rrr|rrr|rrr} 
 \toprule
 & \multicolumn{3}{c|}{\textbf{MUC}} & \multicolumn{3}{c|}{\textbf{B$^{3}$}} & \multicolumn{3}{c|}{\textbf{CEAF$_{e}$}} & \multicolumn{3}{c}{\textbf{BLANC}} \\
 
\textbf{\textbf{Prompt}} & Precision & Recall & F-1 & Precision & Recall & F-1 & Precision & Recall & F-1 & Precision & Recall & F-1\\

 \midrule
 Iterative Prediction & & & & & & & & & & & &\\
 \hspace{.1in}whole doc & $11.5$ & $21.4$ & $15.0$ & $91.3$ & $95.5$ & $93.3$ & $93.0$ & $88.6$ & $90.7$ & $52.5$ & $55.9$ & $53.4$\\ 
 \hspace{.1in}1-shot & $16.7$ & $21.4$ & $18.8$ & $94.3$ & $95.5$ & $94.9$ & $94.2$ & $92.7$ & $93.5$ & $58.9$ & $58.5$ & $58.7$\\ 
 \hspace{.1in}2-shot & $27.3$ & $21.4$ & $24.0$ & $97.0$ & $95.5$ & $96.2$ & $93.5$ & $94.6$ & $94.1$ & $66.4$ & $58.6$ & $61.3$\\ 
 \hspace{.1in}5-shot & $25.0$ & $21.4$ & $23.1$ & $96.6$ & $95.3$ & $96.1$ & $94.0$ & $94.7$ & $94.4$ & $61.5$ & $58.5$ & $59.8$\\ 
 \hspace{.1in}10-shot & $14.3$ & $16.7$ & $15.4$ & $94.6$ & $95.5$ & $95.2$ & $93.7$ & $92.9$ & $93.3$ & $55.3$ & $54.8$ & $55.0$\\
\bottomrule

 \toprule
 & \multicolumn{3}{c|}{\textbf{Temporal}} & \multicolumn{3}{c|}{\textbf{Causal}} & \multicolumn{3}{c|}{\textbf{Subevent}} & \multicolumn{3}{c}{\textbf{Overall}} \\
 
 \textbf{\textbf{Prompt}} & Precision & Recall & F-1 & Precision & Recall & F-1 & Precision & Recall & F-1 & Precision & Recall & F-1\\
 \midrule
  Iterative Prediction & & & & & & & & & & & &\\
 \hspace{.1in}whole doc & $25.8$ & $19.0$ & $21.9$ & $3.2$ & $4.8$ & $3.8$ & $0$ & $0$ & $0$ & $22.8$ & $20.9$ & $22.2$\\
 \hspace{.1in}1-shot & $19.0$ & $10.5$ & $13.6$ & $6.0$ & $6.0$ & $6.0$ & $0$ & $0$ & $0$ & $22.8.$ & $20.9$ & $21.5$\\
 \hspace{.1in}2-shot & $21.1$ & $8.4$ & $12.1$ & $5.4$ & $4.2$ & $4.7$ & $4.3$ & $4.9$ & $4.6$ & $25.5$ & $21.3$ & $22.6$\\
 \hspace{.1in}5-shot & $21.6$ & $11.6$ & $15.1$ & $10.4$ & $6.0$ & $7.6$ & $1.9$ & $2.4$ & $2.1$ & $25.8$ & $21.9$ & $23.3$\\
 \hspace{.1in}10-shot & $25.2$ & $12.7$ & $16.9$ & $12.2$ & $9.4$ & $10.6$ & $0$ & $0$ & $0$ & $25.5$ & $21.8$ & $23.1$\\ 
\bottomrule
\end{tabular}}
\end{center}
\vspace{-0.1in}
\caption{Event coreference, temporal, causal, and subevent resolution performances of \textit{gpt-4-1106-preview} on \mavenere using Iterative Prediction prompt patterns on the first 10 documents of the validation set.}
\label{tab:gpt4-valid-table}
\end{table*}
\begin{table*}[t!]
\small
\begin{center}
\scalebox{0.87}{
\begin{tabular}{ lrrr|rrr|rrr|rrr} 
 \toprule
 & \multicolumn{3}{c|}{\textbf{MUC}} & \multicolumn{3}{c|}{\textbf{B$^{3}$}} & \multicolumn{3}{c|}{\textbf{CEAF$_{e}$}} & \multicolumn{3}{c}{\textbf{BLANC}} \\
\textbf{Prompt} & Precision & Recall & F-1 & Precision & Recall & F-1 & Precision & Recall & F-1 & Precision & Recall & F-1\\
 \midrule
 Iterative Prediction & & & & & & & & & & & &\\
 \hspace{.1in}whole doc & $4.6$ & $7.1$ & $5.6$ & $91.9$ & $94.9$ & $93.3$ & $92.5$ & $90.0$ & $91.0$ & $50.9$ & $51.6$ & $51.1$\\ 
 \hspace{.1in}1-shot & $0$ & $0$ & $0$ & $100.0$ & $94.7$ & $97.3$ & $93.4$ & $98.6$ & $95.9$ & $49.7$ & $50.0$ & $49.9$\\ 
 \hspace{.1in}2-shot & $0$ & $0$ & $0$ & $100.0$ & $94.7$ & $97.3$ & $93.4$ & $98.6$ & $95.9$ & $49.7$ & $50.0$ & $49.9$\\ 
 \hspace{.1in}5-shot & $0$ & $0$ & $0$ & $100.0$ & $94.7$ & $97.3$ & $93.4$ & $98.6$ & $95.9$ & $49.7$ & $50.0$ & $49.9$\\  
\bottomrule
 \toprule
 & \multicolumn{3}{c|}{\textbf{Temporal}} & \multicolumn{3}{c|}{\textbf{Causal}} & \multicolumn{3}{c|}{\textbf{Subevent}} & \multicolumn{3}{c}{\textbf{Overall}} \\
\textbf{Prompt} & Precision & Recall & F-1 & Precision & Recall & F-1 & Precision & Recall & F-1 & Precision & Recall & F-1\\
 \midrule
  Iterative Prediction & & & & & & & & & & & &\\
 \hspace{.1in}whole doc & $16.7$ & $3.5$ & $5.7$ & $5.8$ & $12.5$ & $7.9$ & $3.4$ & $26.8$ & $6.1$ & $22.8$ & $20.9$ & $20.0$\\
 \hspace{.1in}1-shot & $8.4$ & $1.9$ & $3.0$ & $8.7$ & $1.2$ & $2.1$ & $0$ & $0$ & $0$ & $22.8$ & $20.9$ & $16.5$\\
 \hspace{.1in}2-shot & $16.4$ & $3.1$ & $5.2$ & $9.1$ & $0.6$ & $1.1$ & $0$ & $0$ & $0$ & $25.5$ & $21.3$ & $16.8$\\
 \hspace{.1in}5-shot & $13.5$ & $2.7$ & $4.5$ & $0$ & $0$ & $0$ & $0$ & $0$ & $0$ & $25.8$ & $21.9$ & $16.3$\\
\bottomrule
\end{tabular}}
\end{center}
\vspace{-0.1in}
\caption{Event coreference, temporal, causal, and subevent resolution performances of \textit{LLaMA-2} on \texttt{mavenere} using Iterative Prediction prompt patterns on the first 10 documents of the validation set.}
\label{tab:llama-valid-table}
\end{table*}

\subsection{Precision and Recall Results for LLaMA-2 SFT Experiment}
\begin{figure*}[!h]
    \centering
    \begin{subfigure}[b]{0.49\textwidth}
    \centering
    \includegraphics[scale=0.5]{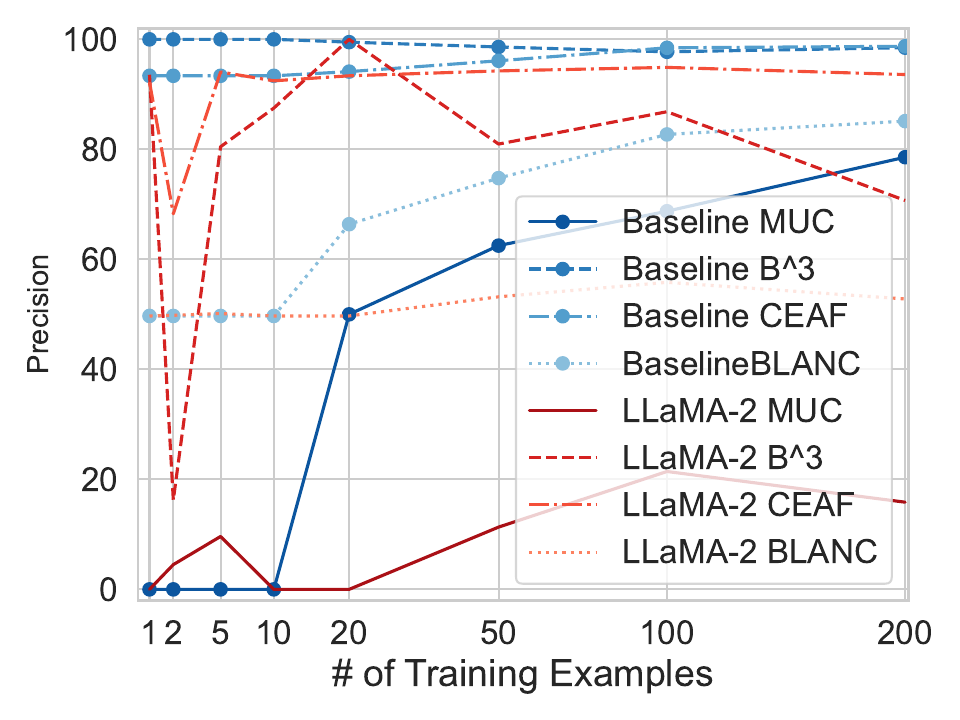}
   \vspace{-0.3cm}
    \caption{Coreference}
    \label{fig:coreference_precision_curve}
    \end{subfigure}
    \hfill
    \begin{subfigure}[b]{0.49\textwidth}
    \centering
    \includegraphics[scale=0.5]{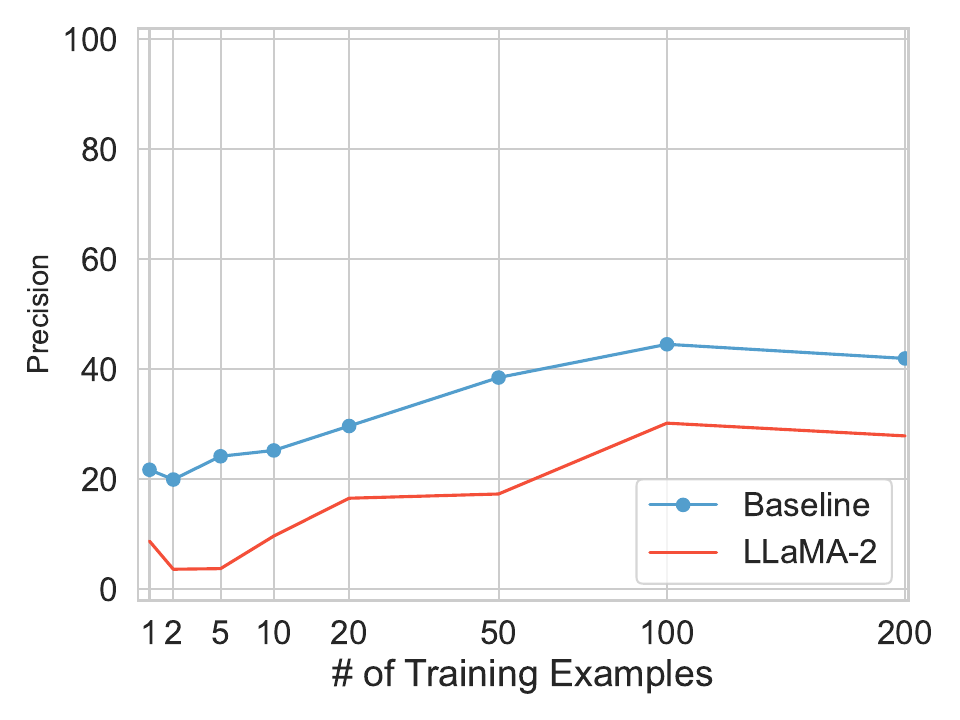}
   \vspace{-0.3cm}
    \caption{Temporal}
    \label{fig:temporal_precision_curve}
    \end{subfigure}
    \hfill
    \begin{subfigure}[b]{0.49\textwidth}
    \centering
    \includegraphics[scale=0.5]{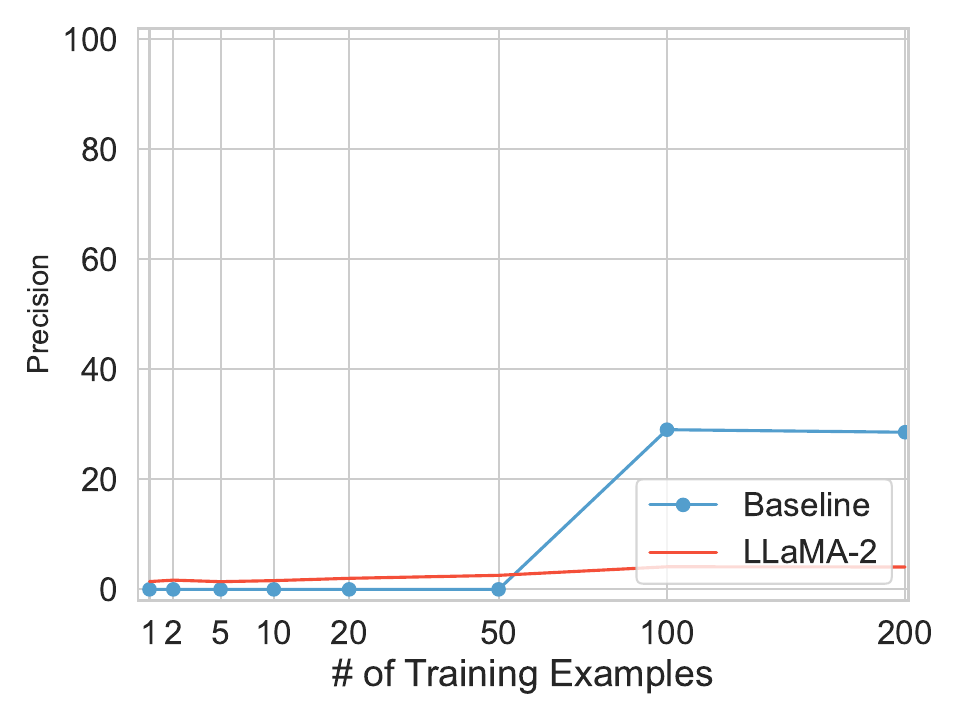}
   \vspace{-0.3cm}
    \caption{Causal}
    \label{fig:causal_precision_curve}
    \end{subfigure}
    \hfill
    \begin{subfigure}[b]{0.49\textwidth}
        \centering
        \includegraphics[scale=0.5]{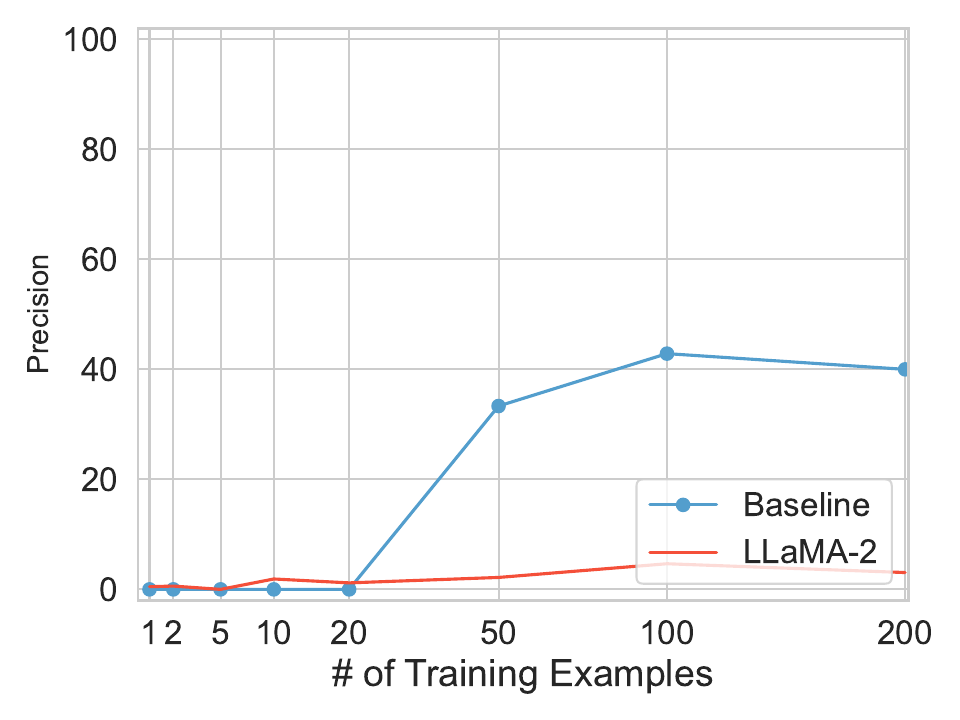}
       \vspace{-0.3cm}
        \caption{Subevent}
        \label{fig:subevent_precision_curve}
    \end{subfigure}
    \caption{Precision scores for coreference, temporal, causal, and subevent relations. The blue lines represent the baseline's performance trained with different numbers of documents. The red lines represent the performance of LLaMA-2 fine-tuned with different numbers of documents.}
    \label{fig:precision-curves}
\end{figure*}

\begin{figure*}[!h]
    \centering
    \begin{subfigure}[b]{0.49\textwidth}
    \centering
    \includegraphics[scale=0.5]{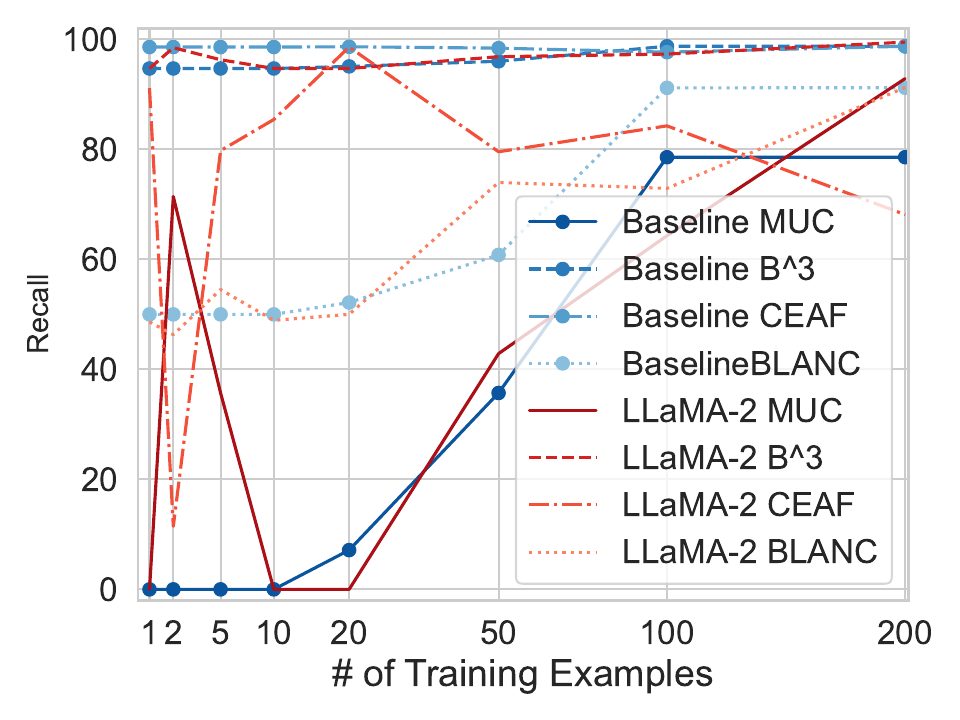}
   \vspace{-0.3cm}
    \caption{Coreference}
    \label{fig:coreference_recall_curve}
    \end{subfigure}
    \hfill
    \begin{subfigure}[b]{0.49\textwidth}
    \centering
    \includegraphics[scale=0.5]{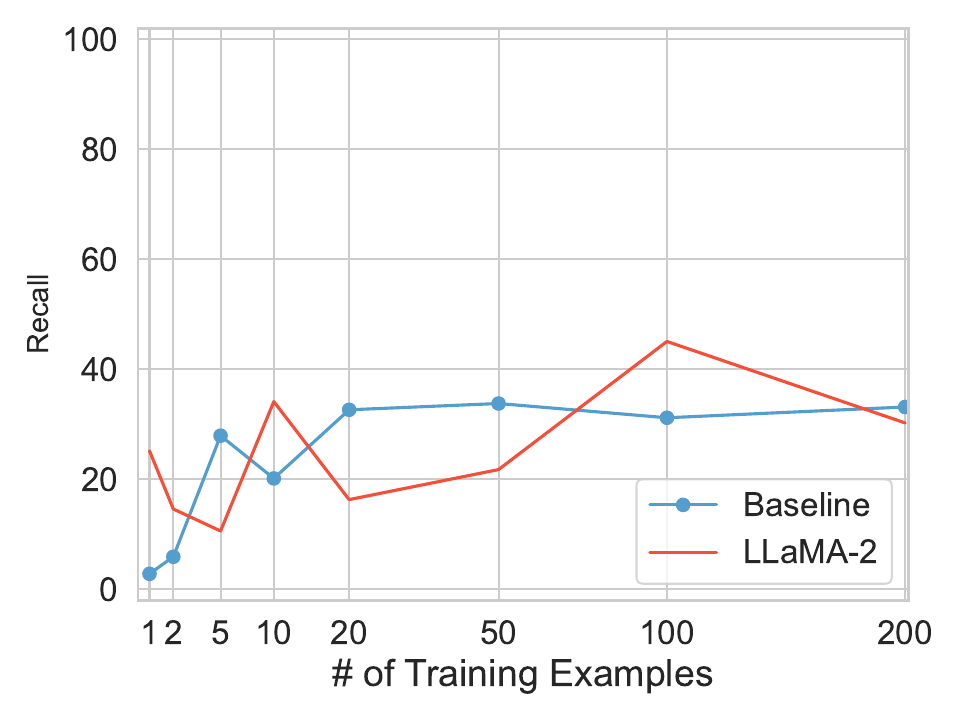}
   \vspace{-0.3cm}
    \caption{Temporal}
    \label{fig:temporal_recall_curve}
    \end{subfigure}
    \hfill
    \begin{subfigure}[b]{0.49\textwidth}
    \centering
    \includegraphics[scale=0.5]{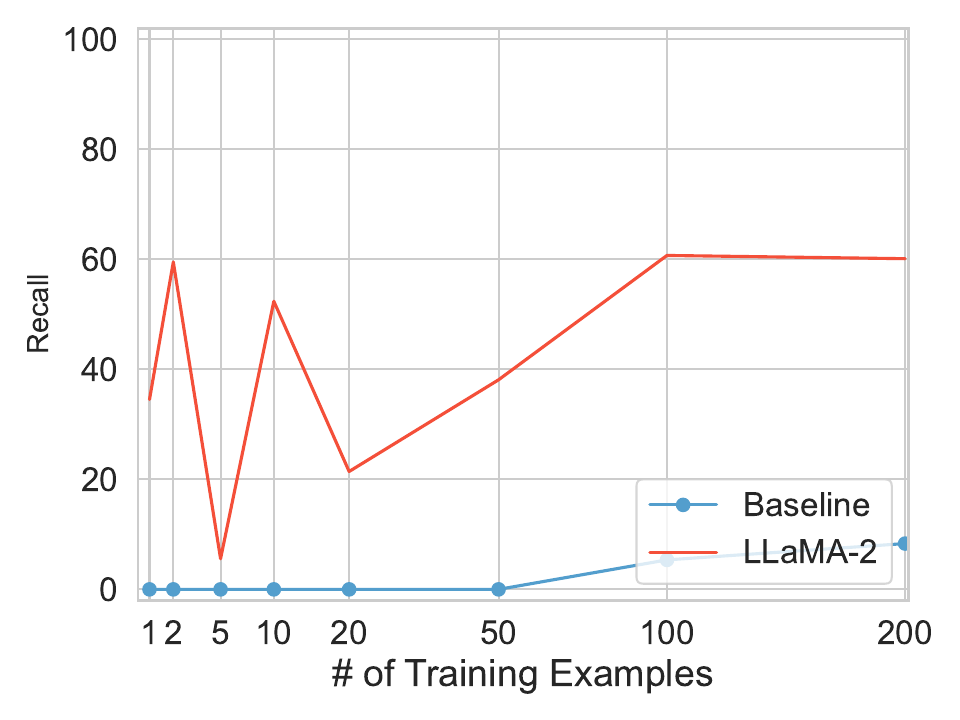}
   \vspace{-0.3cm}
    \caption{Causal}
    \label{fig:causal_recall_curve}
    \end{subfigure}
    \hfill
    \begin{subfigure}[b]{0.49\textwidth}
        \centering
        \includegraphics[scale=0.5]{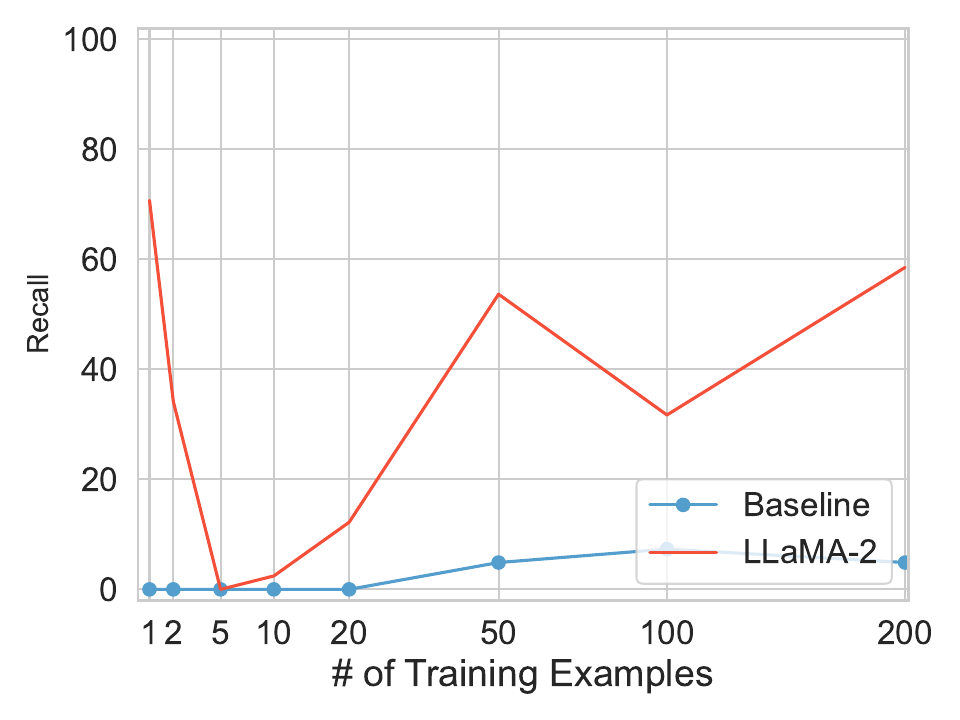}
       \vspace{-0.3cm}
        \caption{Subevent}
        \label{fig:subevent_recall_curve}
    \end{subfigure}
    \caption{Recall scores for coreference, temporal, causal, and subevent relations. The blue lines represent the baseline's performance trained with different numbers of documents. The red lines represent the performance of LLaMA-2 fine-tuned with different numbers of documents.}
    \label{fig:recall-curves}
\end{figure*}

We show the Precision scores of coreference, temporal, causal, and subevent relations for the size experiment discussed in Section~\ref{sec:results} in Figure~\ref{fig:precision-curves}, and we also show the Recall scores of coreference, temporal, causal, and subevent relations for the size experiment discussed in Section~\ref{sec:results} in Figure~\ref{fig:recall-curves}

\label{sec: precision and recall size}

\end{document}